\documentclass[12pt]{article}

\usepackage{scicite}

\usepackage{times}
\usepackage{amsfonts}
\usepackage{siunitx}
\usepackage{algorithm}
\usepackage{algpseudocode}
\usepackage{graphicx}
\usepackage{amsmath, amssymb, amsfonts}
\usepackage{booktabs}
\usepackage[labelformat=empty]{caption}
\usepackage{url}
\usepackage[breaklinks=true]{hyperref}
\usepackage{longtable}
\usepackage{array}

\newcommand{\um}{\,\si{\micro\meter}}

\newenvironment{sciabstract}{%
\begin{quote} \bf}
{\end{quote}}

\title{Robots That Generate Planarity Through Geometry}

\author
{Jakub F. Kowalewski,$^{1*}$ Abdulaziz O. Alrashed,$^{2}$ Jacob Alpert,$^{1}$ Rishi Ponnapalli,$^{3}$ \\ Lucas R. Meza,$^{2}$ and Jeffrey Ian Lipton$^{1\ast}$\\
\\
\normalsize{$^{1}$Department of Mechanical \& Industrial Engineering, Northeastern University,}\\
\normalsize{360 Huntington Avenue Boston, MA 02115, USA}\\
\normalsize{$^{2}$Department of Mechanical Engineering, University of Washington}\\
\normalsize{3900 E Stevens Way NE, WA 98195, USA}\\
\normalsize{$^{3}$Khoury College of Computer Sciences, Northeastern University,}\\
\normalsize{360 Huntington Avenue Boston, MA 02115, USA}\\
\normalsize{\textsuperscript{*}To whom correspondence should be addressed.}\\
\normalsize{
kowalewski.j@northeastern.edu (J.F.K); j.lipton@northeastern.edu (J.I.L)
}}

\date{}

\begin{document} 


\baselineskip24pt


\maketitle

\begin{sciabstract}

Constraining motion to a flat surface is a fundamental requirement for equipment across science and engineering.
Modern precision robotic motion systems, such as gantries, rely on the flatness of components, including guide rails and granite surface plates. However, translating this static flatness into motion requires precise internal alignment and tight-tolerance components that create long, error-sensitive reference chains. 
Here, we show that by using the geometric inversion of a sphere into a plane, we can produce robotic motion systems that derive planarity entirely from link lengths and connectivity. This allows planar motion to emerge from self-referencing geometric constraints, and without external metrology. 
We demonstrate these Flat-Plane Mechanisms (FPMs) from micron to meter scales and show that fabrication errors can be attenuated by an order of magnitude in the resulting flatness. Finally, we present a robotic FPM-based 3-axis positioning system that can be used for metrology surface scans ($\pm \qty{12}{\um}$) and 3D printing inside narrow containers.  
This work establishes an alternative geometric foundation for planar motion that can be realized across size scales and opens new possibilities in metrology, fabrication, and micro-positioning.

\end{sciabstract}

\noindent \textbf{One Sentence Summary} We show that geometric inversion enables scalable, self-referencing planar robots that attenuate fabrication errors and support high-precision metrology and fabrication.


\section*{INTRODUCTION}

\begin{figure}
\makebox[\textwidth][c]{\includegraphics[width=\textwidth]{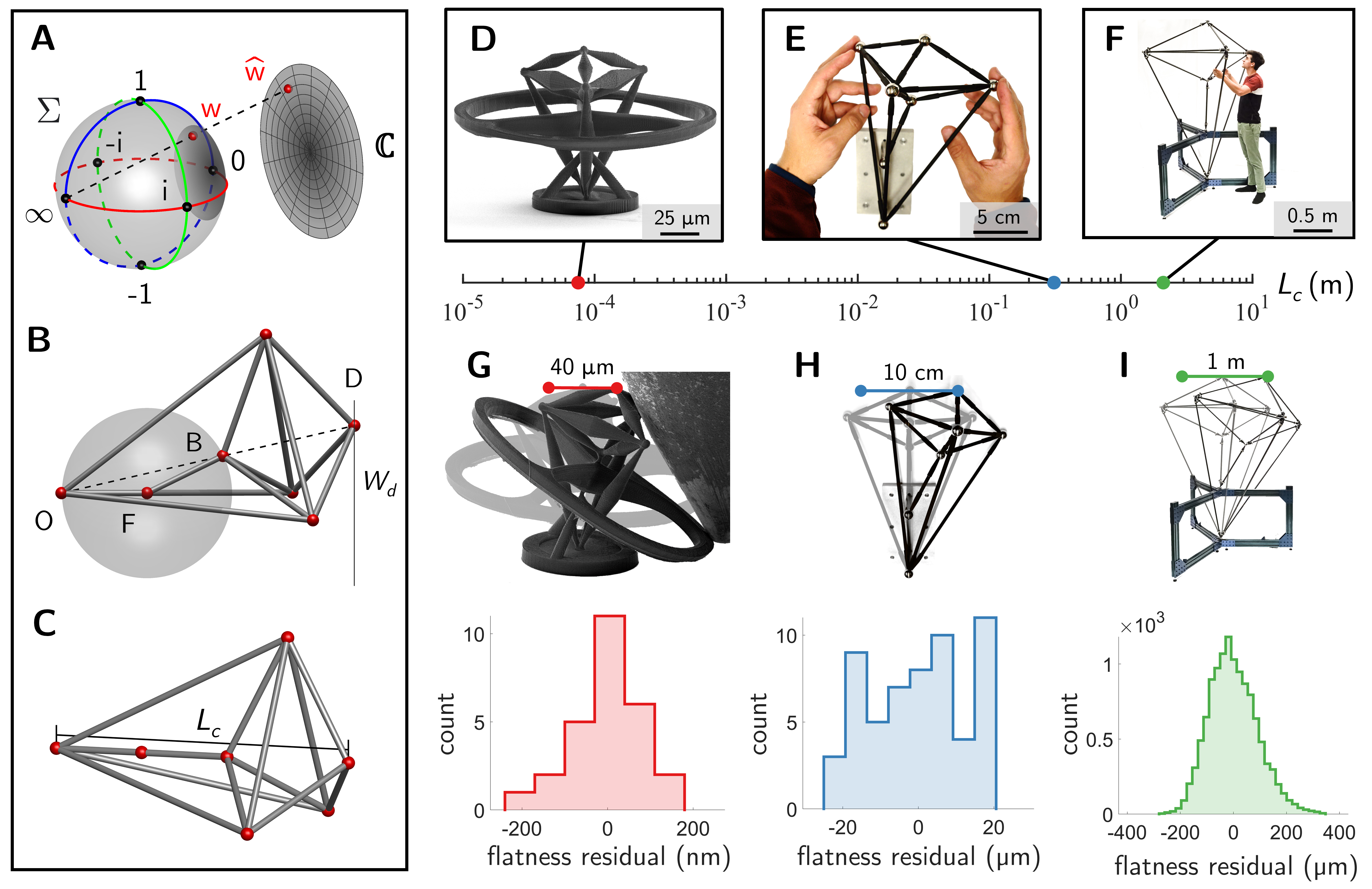}}
\caption{{\bf Fig. 1. The Flat-Plane Mechanism (FPM).} {\bf (A)} Stereographic projection defines the mapping between the Riemann sphere and the complex plane. {\bf (B)} The FPM embodies this mapping in a physical structure, converting spherical input motion into planar output motion. Points $O$ and $F$ are grounded; point $B$ is constrained to a sphere, constraining the endpoint $D$ to a plane. {\bf (C)} The characteristic length, $L_c$, is defined as the tip-to-base distance ($\overline{OD}$) when the mechanism is at the origin ($\overline{OF}$ is collinear with $\overline{FB}$). {\bf (D)} A compliant micro-FPM printed via two-photon lithography. {\bf (E)} A centimeter-scale FPM constructed from wooden dowels and magnetic joints. {\bf (F)} A meter-scale FPM fabricated from carbon fiber rods. {\bf (G to I)} Workspace and flatness measurements for the micro-, centimeter-, and meter-scale FPMs, respectively.}
\label{fig:Mech}
\end{figure}

Modern precision robotic motion systems have enabled broad advances in science and engineering \cite{Winchester2018-ft}. Improving the precision of planar motion has been central to progress in microscopy \cite{Neuman2008-zd,Junker2009-fx,Dufrene2017-cx,Diederich2020-ap}, molecular imaging \cite{Fogliano2021-qu,Alldritt2020-oj,Peng2021-wz}, nanoscale fabrication/patterning \cite{Meza2014-vg,Saha2019-br,Benson2022-hj}, and semiconductor lithography \cite{Schmidt2012-lg,Sreenivasan2017-td,Miyashiro2009-sz}. Traditionally, planar motion is not generated intrinsically but is imposed through extrinsic reference surfaces, calibration, and control \cite{Slocum1992-xu,Castro2003-yi,Li2025-au}. This externalization of precision increases system complexity and constrains portability, scalability, and long-term stability. An alternative is to embed planarity directly within the geometry of a robot itself.

The challenge of maintaining planar motion depends strongly on scale. At the macroscale, planar motion is most commonly produced by Cartesian robots. These robots typically rely on heavy, precisely aligned structures that are difficult to transport and deploy \cite{Stearns2003-ih,Cuypers2009-ly}. A Cartesian robot is constructed from a gantry of independent linear axes arranged in series \cite{Duty2017-vr}. As a result, small angular errors in upstream guide rails propagate and amplify into large positional errors at the tool tip, giving rise to Abbe errors \cite{Slocum1992-xu}. Cable-driven robots avoid serially stacked linear axes and can be realized as deployable structures \cite{Jung2025-qi}, but this comes at the cost of reduced repeatability and tightly constrained workspaces \cite{Jung2025-qi,Atkins2020-vw}. At the microscale, planar motion is typically generated through the elastic deformation of compliant mechanisms to eliminate bearing surfaces and friction \cite{Howell2001-xn}. This approach has enabled a wide range of high-precision flexure stages \cite{Xi2016-wr,Kim2012-dp,Maroufi2015-oj,Olfatnia2013-iu,Li2010-ml,Awtar2013-gs,Lyu2023-ft,Wan2016-ns,Graser2021-dd} as well as miniaturized parallel manipulators \cite{Man2025-lu,Leveziel2022-ex}. However, micro-positioning devices are fundamentally constrained by a trade-off between workspace, precision, and footprint \cite{Lyu2023-ft,Graser2021-dd}. A mechanism that substantially increases workspace relative to footprint while preserving flat motion would offer a new avenue for building compact high-precision micro-positioning devices.


Generating planar motion without referencing manufactured flatness requires a geometry where flat motion is an emergent property. Solving a lower-dimensional version of this problem was critical to the first industrial revolution. Frustrated with the poor linearity of guide rails at the time, Watt invented the first approximate straight-line mechanism (SLM), a device for converting circular motion to precise linear motion \cite{Morley1919-gs}. The Peaucellier-Lipkin mechanism later improved upon Watt's approximation, providing an exact solution for the straight-line motion problem in the 19\textsuperscript{th} century via the geometric inversion of a circle to a line \cite{Kempe1877-gs}. Although it has been theorized that this principle of inversion generalizes to n-dimensional structures \cite{Abbott2008-do}, a physical mechanism that generates flat-planar motion through the geometric inversion of a sphere has yet to be demonstrated.

In this work, we present a solution to the flat-plane mechanism (FPM) problem (Fig.~\ref{fig:Mech}). We demonstrate that our solution is both scalable and robust to fabrication errors. We fabricated FPMs across four orders of magnitude in size scale, from $L_c=\qty{100}{\um}$ to $L_c=\qty{2.12}{\meter}$. At the micro-scale, we show that compliant FPMs are an order of magnitude more compact than existing micro-positioning structures, achieving a workspace-to-footprint ratio of $23.8\%$. At the macro-scale, we show that meter-scale FPMs can be constructed as light-weight, self-aligning, and portable structures. Through a design space analysis, we identify a family of FPM designs that attenuate fabrication errors, and experimentally show that an FPM made of hand-cut wooden dowels produces planar motion with flatness below typical computer numeric control (CNC) machining tolerances of ~$\SI{25}{\micro\meter}$ (Fig.~\ref{fig:Error}). We further demonstrate that FPMs can be fabricated without accurately measuring link lengths, instead bootstrapping the design using an iterative self-comparison process. Finally, we introduce a robotic FPM, capable of scanning surfaces ($\pm \SI{12}{\micro\meter}$) and 3D printing within constrained volumes (Fig.~\ref{fig:actuation}).

\section*{RESULTS}

\subsection*{Geometric scale invariance}
Our FPM produces its planar motion through the geometric inversion of a sphere to a plane (see Supplementary Materials for proof). In three dimensions, this mapping can be represented as the stereographic projection between the sphere $\Sigma$ and complex plane $\mathbb{C}$ (Fig.~\ref{fig:Mech}A) \cite{Needham2021-as, anagnostakis1984arabic}. The FPM is a physical realization of this mapping defined by thirteen links connected at six points labeled in Fig.~\ref{fig:Mech}B. The structure is fixed in place using the ground link, whose length is defined as $\overline{OF}$. The control link $\overline{FB}$ constrains point B to a sphere, and through the properties of geometric inversion, the endpoint D traces a flat disc with diameter $W_d$. If lengths $\overline{OF} \neq \overline{FB}$, the structure inverts the spherical motion of $B$ into spherical motion of point $D$. This embodiment for geometric inversion has no inherent size scale, as it is entirely dependent on the relationships between the points. 

The particulars of any fabrication method introduce a characteristic length $L_c$, which we define as the tip-to-base distance of the mechanism at its origin (Fig.~\ref{fig:Mech}C). We show that FPMs can produce planar motion across four orders of magnitude in size, from $L_c = \SI{100}{\micro\meter}$ to $L_c = \SI{2.12}{\meter}$ (Fig.~\ref{fig:Mech}). For each design, we recorded the output surface by measuring the quasi-static position of the endpoint over a normalized workspace ($W_d/L_c \approx 0.4$). We computed the flatness as the root-mean-square error (RMSE) from a least-squares plane or, when a full plane could not be measured, from a series of best-fit line segments. Across the micro, centimeter, and meter characteristic lengths, we found the root-mean-square (RMS) flatness relative to $L_c$ to be $0.095\%$, $0.005\%$, $0.004\%$, respectively.  

At the centimeter and meter size scales, we fabricated FPMs using rigid links connected with steel spheres and magnets (Fig.~\ref{fig:Mech}E, Movie~S2, Movie~S3). The centimeter-scale FPM achieved high RMS flatness \SI{12.59}{\micro\meter} across a \qty{10}{\cm} workspace. Across the meter-scale FPMs \qty{1}{\meter} diameter workspace, we calculated a RMS flatness of \SI{91.28}{\micro\meter}, close to the measurement noise of our motion-capture system ($\sigma = \SI{72.56}{\micro\meter}$). Both mechanisms are lightweight (\SI{92.88}{\gram} and \SI{4.85}{\kilo\gram}) and can be broken down into rods and spheres for transport, and because the joints self-align, they can be quickly redeployed.

At the micrometer scale, high-precision planar positioning is crucial to semiconductor manufacturing, optics, and microscopy. However, designing micro-scale motion structures that maintain planarity over a large relative workspace is challenging. We fabricated a compliant monolithic micro-FPM ($L_c = \SI{100}{\micro\meter}$) using two-photon lithography (Fig.~\ref{fig:Mech}D). At this size scale, pin and spherical joints are impractical to fabricate, so instead we use compliant flexure joints that approximate the ideal kinematic constraints. We actuated the mechanism by applying a load to the control link through an integrated disk and tracked the endpoint using a scanning electron microscope and determined the resulting RMS flatness to be $\SI{95.33}{\nano\meter}$ (Movie~S1). We show that micro-scale FPMs have a workspace-to-footprint ratio of $W/F = 23.78 \%$, more than an order of magnitude more compact than the current state-of-the-art flexure positioning structures we identified in the literature (Table~S1).

\begin{figure}
\makebox[\textwidth][c]{\includegraphics[width=\textwidth]{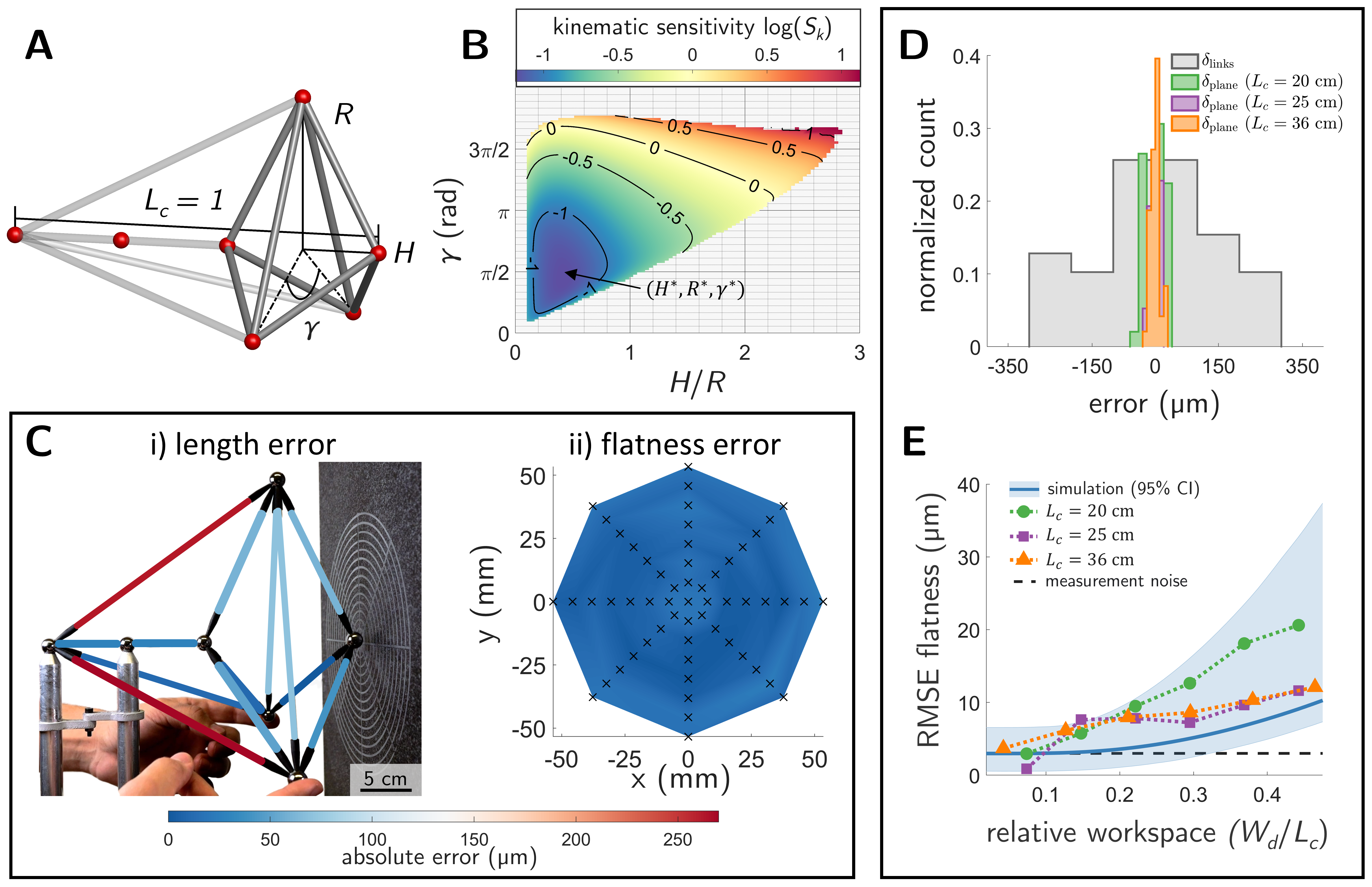}}
\caption{{\bf Fig. 2. Design space and sensitivity analysis.} {\bf (A)} We define the FPM design space using its characteristic length and bi-pyramid dimensions when the mechanism is at the origin. {\bf (B)} The kinematic sensitivity landscape reveals a family of designs that attenuate fabrication errors. {\bf (C)} Mechanisms in this family achieve RMS flatness an order of magnitude lower than the RMSE within their link lengths. {\bf (D)} We experimentally validate these results for FPMs of varying sizes.  {\bf (E)} We also map out the trade-off between workspace size and flatness. Error bars for the simulated data show the 95\% confidence interval from $n=50$ randomized instances.}
\label{fig:Error}
\end{figure}

\subsection*{Error attenuation in the design space}

Physically realizing the FPM inherently introduces fabrication errors in the structure. We conducted a kinematic sensitivity analysis to understand how these fabrication errors propagate to the flatness of an FPM's motion. We first parameterize the FPM design space using the symmetric triangular bipyramid defined by $R$, $H$, and $\gamma$ in the neutral configuration (Fig.~\ref{fig:Error}A). Two geometric error sources contribute to flatness error: link length deviations and joint rotational offsets. We assume that for rigid-bodied FPMs, the dominant source of error is fabrication variance within the structure's link lengths. To quantify how FPM design parameters influence error propagation, we define a dimensionless kinematic sensitivity metric \(S_k\) as the ratio of the RMS flatness to link-length RMSE.

We found that the choice of design parameters strongly influences how fabrication variance propagates to the output surface. In particular, we performed a sensitivity analysis over the FPM's design space and found a global minimum $H^*=1/4,R^*=1/2,\gamma^*=\pi/2$, where $S_{k,\text{min}} = 0.072$. Surrounding this optimum, there exists a large family of designs that achieve RMS flatness an order of magnitude lower than the link-length RMSE (Fig.~\ref{fig:Error}B). To validate these results experimentally, we fabricated three centimeter-scale FPMs with sizes ${L}_c \approx \qty{20}{\cm}$, $Lc \approx \qty{25}{\cm}$, and $L_c \approx \qty{36}{\cm}$ and measured the length and flatness RMSE (Fig.~\ref{fig:Error}C-i). Across these FPMs, the mean link length RMSE was $\SI{141.8}{\micro\meter}$ and the mean flatness RMSE was $\SI{15.6}{\micro\meter}$, demonstrating a close match to the simulation (Fig.~\ref{fig:Error}D). Additionally, we found that since RMS flatness is not normally distributed over the workspace, it can be further improved utilizing a smaller relative workspace (Fig.~\ref{fig:Error}E).

\subsection*{Fabrication without external metrology}

We demonstrate that flat motion can emerge without measurement through an iterative fabrication process for the FPM (Movie~S4). Analogous to the way granite surface plates are flattened by mutual lapping, the FPM improves its own planarity through iterative self-referencing. This process relies only on copying, comparison, and geometric closure operations (see Supplemental Materials). Fabrication is initialized from an arbitrarily chosen seed link, which is copied into imperfect integer-multiple lengths. We identify a set of polygonal loops constructed from relationships of the link copies that are only closable when the link lengths approximate the design optimum $H^*, R^*, \gamma^*$ (Fig.~\ref{fig:msgd}). By testing loop closure, we generate a comparison-based feedback signal that guides iterative updates to the link lengths and progressively improves planarity. This process removes the need for absolute length measurements of FPM's links.

Each iteration of testing loop-closure can be used to produce a functional mechanism.  Starting from a poorly performing mechanism with $S_k = 0.392$, a single refinement step improves the kinematic sensitivity by more than fivefold to $S_k \approx 0.077$ (Fig.~S\ref{fig:msgd}C). With two additional iterations, the $S_k$ only gradually improves to $S_k=0.073$. This is consistent with having converged to the stable basin of design around the optimum in Fig.~\ref{fig:Error}B. Together, these results show that flat motion can be bootstrapped through a small number of comparison-based fabrication steps, without any reliance on absolute dimensional control.

\begin{figure}
\makebox[\textwidth][c]{\includegraphics[width=0.95\textwidth]{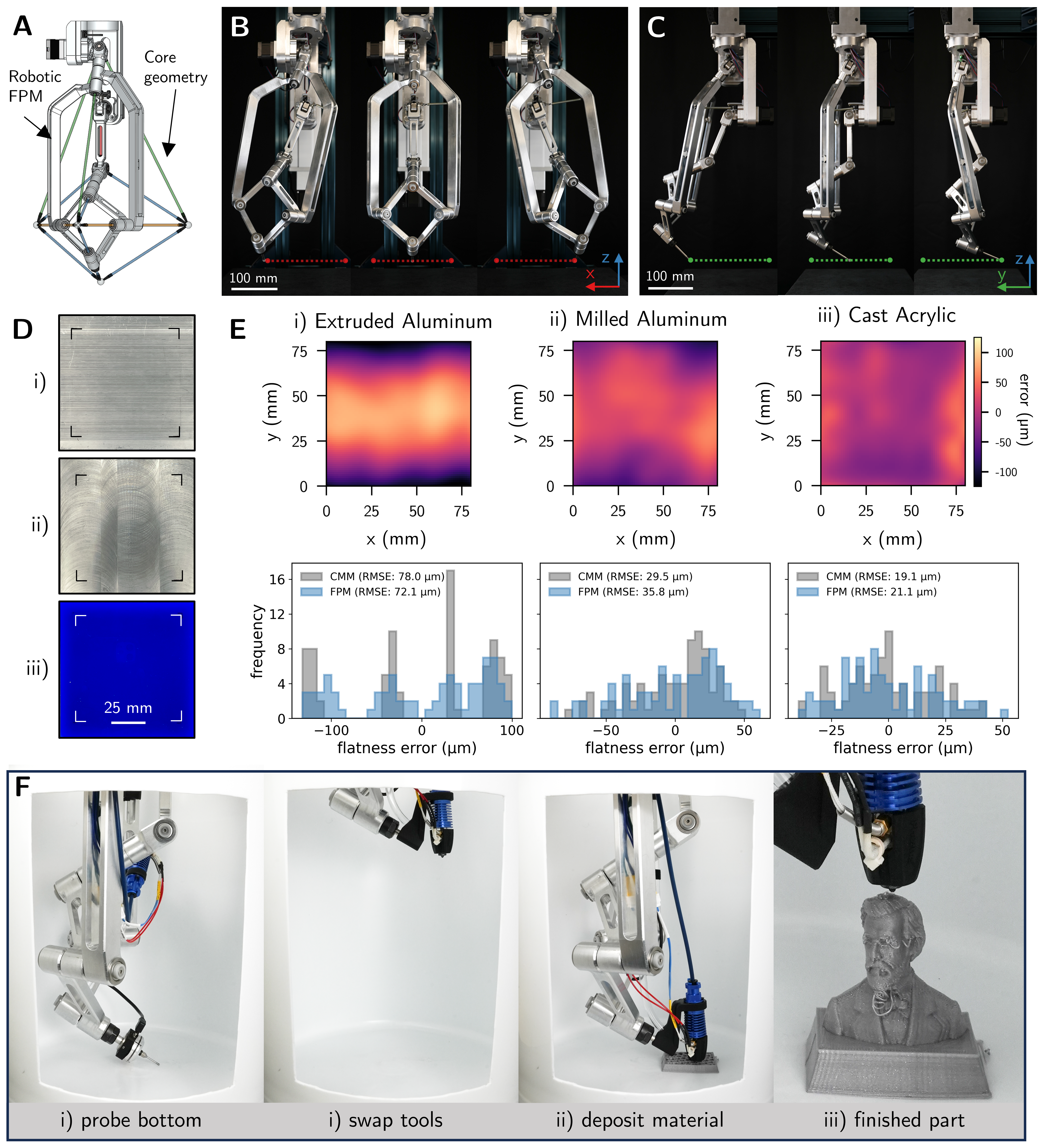}}
\caption{{\bf Fig. 3. A robotic FPM for metrology and fabrication.} {\bf (A)} The actuated FPM, overlaid with the fixed-distance constraints. This FPM's workspace spans \qty{200}{\mm} in the X {\bf (B)} and Y {\bf (C)} directions. {\bf (D)} Surfaces scanned using the FPM: extruded aluminum, milled aluminum, and cast acrylic. {\bf (E)} Comparison of the FPM's scans with a lab-grade CMM. {\bf (F)} The robot scanning the bottom of a narrow container and then 3D printing a model of Yom Tov Lipkin, one of the inventors of the first exact straight-line mechanism.}
\label{fig:actuation}
\end{figure}

\subsection*{Demonstrating an FPM-based robot}

To demonstrate how Flat-Plane Mechanisms make precision planar motion more accessible to robotics, we developed a demonstration system and applied it to metrology and fabrication tasks (Fig.~\ref{fig:actuation}). Rather than relying on closed-loop sensing, sensitive component alignment, or specialized actuators, the robot generates high-precision planar motion directly through its kinematic structure. The mechanism is composed entirely of single–degree-of-freedom revolute joints, enabling direct actuation using only two standard harmonic drive motors.

This design removes spherical joints entirely by decomposing the FPM's core geometry into six triangles (Fig.~S\ref{fig:triangles}). In our robot's design, each triangle is enforced through a fixed angle constraint, which we realize using machined aluminum links (Fig.~\ref{fig:actuation}A). The ground link houses both harmonic drives, which drive the spherical motion of the control link that is then mapped into the planar motion of the endpoint. A derivation of the inverse kinematics is provided in Supplemental Materials. We attached the FPM to a linear rail for positioning in the $Z$ direction.

The robotic FPM has a characteristic length of $L_c = \qty{487.5}{\mm}$, a planar workspace of $W_d = \qty{200}{\mm}$, and a vertical ($Z$) travel range of $\qty{300}{\mm}$ (Fig.~\ref{fig:actuation}B,C). Despite its minimal actuation and mechanical complexity, the system achieves high-precision planar motion, with single-point repeatability of $\sigma_x = \qty{1.9}{\um}$ and $\sigma_y = \qty{2.6}{\um}$ in the $X$ and $Y$ directions, respectively (Fig.~S\ref{fig:xy_repeatability}). By mounting a touch probe at the end effector, we scanned a precision granite surface plate and measured an RMSE planarity of $\qty{62.5}{\um}$, with a $Z$-axis repeatability of $\sigma = \qty{6.0}{\um}$. After compensating for constant geometric errors, residual tool-position errors are dominated by repeatability rather than systematic bias.

We demonstrated that the robotic FPM can be used as a coordinate measurement machine (CMM). To benchmark the system, we scanned three $80\times\qty{80}{\mm}$ flat samples: extruded aluminum (EA), milled aluminum (MA), and cast acrylic (CA) (Fig.~\ref{fig:actuation}D and Movie~S5). We compared these scans against measurements from a laboratory-grade CMM (Hexagon 7.10.7 SF). The RMSE values for the FPM and CMM were \qty{72.1}{\um} and \qty{78}{\um} for EA, \qty{35.8}{\um} and \qty{29.5}{\um} for MA, and \qty{21.1}{\um} and \qty{19.1}{\um} for CA, with a mean difference of \qty{7.1}{\um} (Fig.~\ref{fig:actuation}E). Additional details on the setup and calibration of our mechanism are provided in Materials and Methods. These results show close agreement between the FPM and the gantry-style CMM, highlighting the FPM’s potential in precision metrology.

To demonstrate the utility in fabrication, we integrated a fused filament fabrication (FFF) printhead with the robotic FPM to produce a 3D printer that can reach into narrow spaces (Fig.~\ref{fig:actuation}F and Movie~S6). We first scanned a $\qty{40}{\mm} \times \qty{60}{\mm}$ grid on the bottom of a narrow container (\qty{260}{\mm} diameter and \qty{360}{\mm} depth) to align the FPM's plane to the container's surface and compensate for any warping. We then generated a 3D model of Yom Tom Lipkin, one of the inventors behind the Peaucellier–Lipkin Mechanism, from a 2D image of him \cite{NLI_LipkinArchive}, which we then printed using the FPM, a 3D generalization of his original 2D SLM. This demonstration highlights the FPM’s utility in both metrology and fabrication within constrained environments.

\section*{DISCUSSION}

We demonstrated that Flat-Plane Mechanisms (FPMs) provide a basis for constructing motion systems that derive planarity from self-referencing geometric constraints rather than from long, error-sensitive reference chains tied to static flatness. This principle, based on the geometric inversion of a sphere into a plane, operates independently of fabrication scale. We fabricated FPMs spanning four orders of magnitude, from the micro-scale to the meter-scale. At the micro-scale, FPMs achieved an order-of-magnitude improvement in workspace-to-footprint ratio relative to state-of-the-art flexure stages, suggesting a path toward higher-density positioning architectures for semiconductor manufacturing and microscopic imaging. At the meter scale, the FPM’s inherent self-alignment and portability offer potential advantages in large-format additive manufacturing and field-deployable metrology systems.

We also showed that FPMs can generate flat motion with tolerances that exceed the precision of their fabricated components. Our kinematic sensitivity analysis identified a family of designs that attenuate fabrication errors by approximately an order of magnitude. We validated this behavior experimentally by achieving sub-\qty{25}{\um} flatness using an FPM constructed from hand-cut wooden dowels. Such tolerances are typically associated with CNC-machined systems and highlight the FPM’s potential to democratize precision by enabling high-performance positioning devices to be built using accessible, low-precision fabrication methods. In addition, we found that flatness improves as the relative workspace is reduced, which could enable an FPM to function as both a coarse positioner over a large range and a precision stage within a smaller workspace.

Finally, we designed a robotic three-dimensional positioning system based on an FPM and demonstrated its use in metrology and fabrication. Unlike conventional gantry-based systems, the FPM does not rely on a surrounding structural frame, enabling surface scanning and additive manufacturing within confined volumes. In metrology experiments, FPM-based surface scans closely matched measurements from a laboratory-grade coordinate measuring machine, with deviations within \qty{7.1}{\um}, using a system with a bill of materials totaling \$7{,}281.27 (Table~S3). This will increase the accessibility and deployability of metrology systems. The specific FPM geometry used here prioritizes planarity and reach over lateral stiffness and is therefore best suited for low-contact-force tasks such as probing, scanning, and material deposition. Applications requiring sustained high forces, including CNC milling or heavy manipulation, remain better served by conventional Cartesian architectures. Similar to how multiple straight-line mechanism variants exist, we have only explored a single flat-plane mechanism variant within a broader design space, with alternative geometries and trade-offs in workspace, stiffness, and actuation complexity.

\section*{MATERIALS AND METHODS}

\subsection*{Design, fabrication, and characterization}

\subsubsection*{Micro-scale FPM}

We fabricated the micro-scale FPM (Fig.~1D) using two-photon lithography (Nanoscribe Photonic Professional GT) with IP-Dip photoresist. We used the Galvo scan mode (layer-by-layer printing) where an STL CAD file was sliced with the following writing parameters: Slicing Distance $=\qty{300}{\nm}$, Hatching Distance $=\qty{200}{nm}$, Scanning speed $=10$-mm/sec, Laser power $=19$-mW. The structure was printed with thin supports under the overhang sections to eliminate any misalignments during the printing process. We placed the sample inside a downstream oxygen plasma etcher (YES Asher CV200 RFS) to remove these supports. All link lengths and design parameters are included in Table~S1.

We manipulated the structure using a displacement-controlled piezo-driven nano-indentation system (ASA, Alemnis AG) placed inside a scanning electron microscope (TFS Apreo variable-pressure SEM). The indenter tip actuated the disk connected around the structure following a $\qty{2}{\um}$ displacement increment. Both videos and images were taken for each increment to track the endpoint of the structure.

\subsubsection*{Centimeter-scale FPM}

We fabricated the centimeter-scale FPM’s links (Fig.~1E) using $3/16$~in. wooden Birch rods. To create multiple links of the same length, we clamped stock material in a vice and adjusted the exposed length until it closely matched a previously cut reference link using tactile feedback for comparison. The wooden dowels were cut with a thin-bladed handsaw and the edges were lightly sanded to remove burrs. We used $9/16$~in. diameter 52100 alloy steel spheres for the center of each joint. We connected the links to the steel joint using plastic caps printed from Urethane Methacrylate on a Carbon M1 3D printer with embedded $1/8$~in. diameter Neodymium magnets. The printed end caps were fixed to the links using cyanoacrylate. The joints were cleaned with WD-40, and the magnets were cleaned of debris using tape. For link length measurements, we used 12~in. Pittsburgh Digital Calipers with a measurement accuracy of $\pm \qty{100}{\um}$. To fixture the ground link, we turned two $3/4$~in. aluminum rods on a lathe and tapered the ends to hold the ground joints. We tapped the bottom of the aluminum rods and attached them to a slotted aluminum plate cut on a waterjet. The distance between the joints was set using a wooden ground link before tightening each rod into place and removing the wooden link.

We measured the output surface against a Standridge Laboratory Grade AA Granite Plate with a flatness tolerance of $\pm\,\qty{0.127}{\um}$. We measured the distance between the steel sphere endpoint and granite slab using a set of MG Grade~B Gauge Blocks with a $\pm\,\qty{1.27}{\um}$) tolerance. We determined the repeatability of our measurements to be $\sigma = \qty{3.1}{\um}$, based on $n=15$ measurements of the endpoint's position in the home state.

\subsubsection*{Meter-scale FPM}
We fabricated the meter-scale FPM’s links (Fig.~1F) using $3/4$~in. carbon fiber rods cut using a composite saw. The end caps were 3D printed from Acrylonitrile Butadiene Styrene (ABS) with embedded Neodymium magnets and were attached to the carbon-fiber links using a hose clamp. The links were joined using $1\tfrac{1}{2}$~in. 52100 Alloy steel spheres. The ground link support structure was made from an extruded aluminum frame. The frame had a lockable degree of freedom that allowed the top joint to move in the vertical direction. We used this to set the distance with a carbon fiber ground link, which was then removed before testing.

To measure the output surface, we used an OptiTrack motion capture system with twelve PrimeX~22 cameras. We attached six retro-reflective markers to the end sphere and recorded the FPM’s motion across a trajectory that evenly covered the workspace. For each position, we fit a least-squares sphere to the six markers and used its center as the tracked endpoint. To estimate the OptiTrack’s measurement noise, we recorded the radial distance of each marker on the sphere over the entire trajectory. We found the standard deviation to be $\pm \qty{72.56}{\um}$, which we used as the measurement noise threshold for the system.

\subsubsection*{Robotic FPM}

The robotic FPM was constructed from machined 6061 aluminum links connected by single-degree-of-freedom revolute joints. The robot uses two NEMA-17 stepper motors coupled to 50:1 harmonic drive gearboxes (PGFUN) for actuation. We mounted the mechanism to a vertical linear axis (RATTMMOTOR FBX-500mm) to enable volumetric motion. A tension spring was installed between the ground link and the moving structure to preload the joints and reduce backlash. A full bill of materials is provided in Table~S3. Motion control was implemented on an Octopus V1.1 controller running a custom-modified version of Marlin~2.1.2.4. The firmware included a C++ implementation of the FPM inverse kinematics (Supplementary Materials), allowing G-Code files to be converted directly to motion commands. Homing was performed using optical endstops (MakerHawk). The machine's origin was defined by jogging the FPM until the control link was parallel to the ground link, verified using gauge blocks, and recording the corresponding offsets in firmware.

Tools were mounted at the FPM end-effector using an ER11 spring collet. Surface measurements were performed using a PGFUN TR-0041\_NO touch probe ($\pm \qty{10}{\um}$ repeatability), while additive fabrication used a Phaetus Rapido~2 hot end with a \qty{0.4}{\mm} nozzle. To ensure coincidence between the tool tip and the FPM kinematic end-point, tool offsets were measured using calipers, and a corresponding spacer was inserted. For the surface probing results (Fig.~\ref{fig:actuation}E), the location of each sample's edges was first determined in order to locate the sample relative to the FPM's coordinate frame. Surface scans were then performed using a $9\times9$ probing grid inset by \qty{10}{\mm} from the sample edges. All surface data were analyzed in a sample-relative coordinate frame. 

The robot 3D printed the Yom Tov Lipkin model (Fig.~\ref{fig:actuation}F) using 1.75 mm polylactic acid (PLA) filament and a 0.4 mm nozzle. The printed model had overall dimensions of $\qty{37.96}{\mm} \times \qty{57.30}{\mm} \times \qty{51.37}{\mm}$ and was fabricated using $20\%$ infill, perimeter speeds of \qty{60}{\mm\per\s}, and infill speeds of \qty{80}{\mm\per\s}. The total print time was \qty{2}{h} \qty{58}{min}. Prior to printing within confined geometries, a local planar scan was performed to align the FPM plane to the target surface and compensate for warping.

\subsection*{Sensitivity analysis}

We define the kinematic sensitivity for an FPM as,
\begin{equation}
S_k = \frac{\sqrt{\sum_{i=1}^{n} \frac{(\delta_{\text{plane}})^{2}}{n}}}{\sqrt{\sum_{j=1}^{13} \frac{(\delta_{\text{link}})^{2}}{13}}}
\end{equation}
where \(\delta_{\text{link}}\) are the length deviations of the FPM's thirteen links and \(\delta_{\text{plane}}\) are the residuals of $n$ measured points from the best-fit plane. Our sensitivity analysis (Fig.~\ref{fig:Error}B) evaluated the sensitivity of one million FPM designs over a normalized design space ($L_c=1$). For each design, we generated fifty instances with synthetic fabrication $\sigma = 0.05\%\,L_c$ in the link lengths. We then computed an average $S_k$ from $n=50$ points within a relative workspace size $W_d/L_c = 0.4$ using the forward kinematics (see Supplementary Materials). We then mapped out the performance landscape by compressing the design space from $\mathbb{R}^3 \rightarrow \mathbb{R}^2$ using the ratio of $H/R$.

\subsection*{Calibration}

To remove repeatable, position-dependent vertical bias in FPM measurements, we estimate and subtract a spatially varying Z-error field derived from a reference scan of our granite surface plate (Fig.~S\ref{fig:granite}). Let the calibration scan provide samples $\{(x_i,y_i,z_i)\}_{i=1}^N$ expressed in the FPM coordinate frame. We first remove arbitrary tilt and offset by fitting a best-fit plane
\begin{equation}
z \approx a x + b y + c,
\end{equation}
where the coefficients $(a,b,c)$ are obtained via least-squares minimization. The residuals
\begin{equation}
e_i = z_i - (a x_i + b y_i + c)
\end{equation}
define a repeatable Z-error field over the scanned domain, which we attribute to systematic FPM and probe effects rather than granite flatness.

To evaluate the Z error at arbitrary query locations, we construct a bilinear interpolant $\hat e(x,y)$ from the residual samples $\{(x_i,y_i,e_i)\}$. The resulting model provides a continuous estimate of the Z error within the convex hull of the calibration scan, with out-of-domain queries handled by boundary clipping or omission. For a subsequent surface scan $\{(x_m,y_m,z_m)\}_{m=1}^M$ acquired over the same XY region, compensation is applied pointwise as
\begin{equation}
z_m^{\mathrm{corr}} = z_m - \hat e(x_m,y_m),
\end{equation}
which removes any constant Z bias.

\bibliography{scibib}

\bibliographystyle{Science}

\section*{Acknowledgments}

\textbf{Funding}: This work was funded by NSF grants 2017927, 2212049, and 2035717 by ONR grant 304220, and through the Murdock Charitable Trust through Grant No. 201913596. \textbf{Patents:} U.S. Patent No. 12,181,025, assigned to the University of Washington, covers aspects of this work. \textbf{Availability of data and materials:} All data and materials will be publicly available online.


\section*{Supplementary materials}
Materials and Methods\\
Supplementary Text\\
Figs. S1 to S11\\
Tables S1 to S3\\


\clearpage

\clearpage

\pagenumbering{gobble}


\begin{center}
    \vspace*{\stretch{0.25}} 
    \LARGE{Supplementary Materials for}
    \\
    \vspace{0.5cm}
    \textbf{\Large{Robots That Generate Planarity Through Geometry}} 
    \vspace{1cm} 
    \\
    \large{Jakub F. Kowalewski,$^{1\ast}$ Abdulaziz O. Alrashed,$^{2}$ Jacob Alpert,$^{1}$ \\ Rishi Ponnapalli,$^{3}$  Lucas R. Meza,$^{2}$ and Jeffrey Ian Lipton$^{1\ast}$} \\
    \vspace{0.5cm}
    \normalsize{$^{1}$Department of Mechanical \& Industrial Engineering, Northeastern University,}\\
    \normalsize{360 Huntington Avenue Boston, MA 02115, USA}\\
    \normalsize{$^{2}$Department of Mechanical Engineering, University of Washington}\\
    \normalsize{3900 E Stevens Way NE, WA 98195, USA}\\
    \normalsize{$^{3}$Khoury College of Computer Sciences, Northeastern University,}\\
    \normalsize{360 Huntington Avenue Boston, MA 02115, USA}\\
    \vspace{0.5cm}
    \normalsize{$^\ast$To whom correspondence should be addressed; E-mail:} \normalsize{kowalewski.j@northeastern.edu (J.F.K); j.lipton@northeastern.edu (J.I.L)} \\
    \vspace*{\stretch{0.5}} 
\end{center}

\noindent \textbf{The PDF file includes:} \\
Supplementary Methods \\
Figs. S1 to S11 \\

\noindent \textbf{Other Supplementary Material for this manuscript includes the following:} \\
Movies S1 to S6

\newpage

\section*{Supplementary Text}

\subsection*{Flat-Plane Mechanism proof}
We prove that the FPM (Fig.~S1) constrains a point to move on a plane using three steps. First, we prove the collinearity of $\mathbf{O}$, $\mathbf{B}$, and $\mathbf{D}$. Next, we show that the length of $\overline{\mathbf{O}\mathbf{B}}$ times the length of $\overline{\mathbf{O}\mathbf{D}}$ is equal to a constant. Then we show that $\mathbf{B}$ moves along a sphere that intersects $\mathbf{O}$ causing the inversion of the sphere into a plane.

\paragraph{1)} The points $\mathbf{A}$, $\mathbf{E}$, and $\mathbf{C}$ are on the surface of the sphere $S_{\mathbf{O}}$ centered on point $\mathbf{O}$. This is physically manifested using links of equal length that connect $\mathbf{A}$, $\mathbf{E}$, and $\mathbf{C}$ to point $\mathbf{O}$. Points $\mathbf{A}$, $\mathbf{E}$, and $\mathbf{C}$ also form the plane $P$ that intersects $S_{\mathbf{O}}$, so points $\mathbf{A}$, $\mathbf{E}$, and $\mathbf{C}$ form a circle on plane $P$. We construct points $\mathbf{B}$ and $\mathbf{D}$ on opposite sides of $P$ by having links of the same length connect the point $\mathbf{B}$ and $\mathbf{D}$ to the points $\mathbf{A}$, $\mathbf{E}$, and $\mathbf{C}$ so $\overline{\mathbf{B}\mathbf{G}}$ and $\overline{\mathbf{G}\mathbf{D}}$ are equal and are therefore mirrors of each other through the plane $P$ so the line $\overline{\mathbf{B}\mathbf{D}}$ must be normal to $P$, and intersect $P$ at point $\mathbf{G}$, and $\mathbf{B}$ and $\mathbf{D}$ projected onto the plane are the point $\mathbf{G}$. Then the point $\mathbf{G}$ must be the center of the circle containing $\mathbf{A}$, $\mathbf{E}$ and $\mathbf{C}$ in the plane so the lengths of $\overline{\mathbf{A}\mathbf{G}}$, $\overline{\mathbf{C}\mathbf{G}}$, and $\overline{\mathbf{E}\mathbf{G}}$ must be equal. The line $\overline{\mathbf{O}\mathbf{D}}$ passes through plane $P$. Because $\mathbf{O}$ is equidistant to $\mathbf{A}$, $\mathbf{E}$ and $\mathbf{C}$ the projection of $\mathbf{O}$ onto $P$ must be the point $\mathbf{G}$. Since $\mathbf{O}$, $\mathbf{B}$, and $\mathbf{D}$ all project onto the point $\mathbf{G}$, then $\mathbf{O}$, $\mathbf{B}$ and $\mathbf{D}$ must be colinear.

\paragraph{2)} Triangles $\triangle \mathbf{B}\mathbf{E}\mathbf{G}$ and $\triangle \mathbf{G}\mathbf{E}\mathbf{D}$ are congruent since $\overline{\mathbf{B}\mathbf{E}}\cong\overline{\mathbf{E}\mathbf{D}}$, $\overline{\mathbf{B}\mathbf{G}}\cong\overline{\mathbf{G}\mathbf{D}}$, and the shared side $\overline{\mathbf{G}\mathbf{E}}$ is congruent to itself. Thus, $\angle \mathbf{B}\mathbf{G}\mathbf{E}=\angle \mathbf{D}\mathbf{G}\mathbf{E}$. Since $\angle \mathbf{B}\mathbf{G}\mathbf{E} + \angle \mathbf{D}\mathbf{G}\mathbf{E} = 180^\circ$, then $2\cdot\angle \mathbf{B}\mathbf{G}\mathbf{E}=180^\circ$ and $2\cdot\angle \mathbf{D}\mathbf{G}\mathbf{E}=180^\circ$. Therefore $\angle \mathbf{B}\mathbf{G}\mathbf{E}=90^\circ$ and $\angle \mathbf{D}\mathbf{G}\mathbf{E}=90^\circ$. Let $h=\overline{\mathbf{B}\mathbf{G}}=\overline{\mathbf{G}\mathbf{D}}$, $r=\overline{\mathbf{E}\mathbf{G}}=\overline{\mathbf{C}\mathbf{G}}=\overline{\mathbf{E}\mathbf{G}}$, and $\ell=\overline{\mathbf{O}\mathbf{B}}$. Then
\begin{align}
\overline{\mathbf{O}\mathbf{B}}\cdot\overline{\mathbf{O}\mathbf{D}}&=\ell(\ell+2h)=\ell^2+2h\ell,\\
\overline{\mathbf{O}\mathbf{E}}^2&=(\ell+h)^2+r^2=\ell^2+2\ell h + h^2 + r^2,\\
\overline{\mathbf{E}\mathbf{D}}^2&=h^2+r^2,\\
\Rightarrow\; \overline{\mathbf{O}\mathbf{E}}^2-\overline{\mathbf{E}\mathbf{D}}^2&=\ell^2+2h\ell=\overline{\mathbf{O}\mathbf{B}}\cdot\overline{\mathbf{O}\mathbf{D}}.
\end{align}
We know $\overline{\mathbf{O}\mathbf{E}}$ and $\overline{\mathbf{E}\mathbf{D}}$ are fixed lengths. Thus, $\overline{\mathbf{O}\mathbf{B}}\cdot\overline{\mathbf{O}\mathbf{D}}=k^2$. Since points $\mathbf{O}$, $\mathbf{B}$, and $\mathbf{D}$ are colinear, $\mathbf{D}$ is the inverse of $\mathbf{B}$ relative to sphere $S_{\mathbf{O}}$ centered at $\mathbf{O}$ with radius $k^2$.

\paragraph{3)} By the properties of inverse geometry, if $\mathbf{B}$ is bound to a sphere that contains $\mathbf{O}$, then the sphere will invert into a plane. We construct the sphere $S_{\mathbf{F}}$ by making two links of the same length that connect the points $\mathbf{O}$ and $\mathbf{B}$ to $\mathbf{F}$. By holding the link $\overline{\mathbf{O}\mathbf{F}}$ fixed, we constrain $\mathbf{B}$ to move on sphere $S_{\mathbf{F}}$ which contains $\mathbf{O}$ and therefore the point $\mathbf{D}$ must move along a plane.

\paragraph*{Note: Inversion between two spheres.}

If the distances $\overline{\mathbf{O}\mathbf{F}}$ and $\overline{\mathbf{F}\mathbf{B}}$ are unequal, then $\mathbf{B}$ instead traces a sphere that does not pass through $\mathbf{O}$. Under inversion about $S_{\mathbf{O}}$, such a sphere maps to another sphere rather than to a plane. Consequently, the mechanism constrains motion to a curved surface, with the curvature set by the ratio of these two link lengths rather than by the overall scale of the mechanism.

\subsection*{FPM Design Parameters}
The mechanism is composed of thirteen links connected with six joints. In total, there are four distinct sets of links with equal length, $A=\{A_0,A_1\}=\{\overline{OF},\,\overline{FB}\}$, $B=\{B_0,\dots,B_5\}=\{\overline{BC},\,\overline{BA},\,\overline{BE},\,\overline{CD},\,\overline{AD},\,\overline{ED}\}$, $C=\{C_0,C_1,C_2\}=\{\overline{OC},\,\overline{OA},\,\overline{OE}\}$, and $D=\{D_0,D_1\}=\{\overline{AE},\,\overline{AC}\}$. The mapping from FPM design parameters to link lengths: $A = (L_c-2H)/2$, $B=\sqrt{H^2+R^2}$, $C=\sqrt{(L_c-H)^2+R^2}$, and $D=\sqrt{(R+R\cos(\gamma/2))^2+(R\sin(\gamma/2))^2}$.

\subsection*{Design space boundaries}
The space of valid FPM designs is bounded by four different singular configurations. These singularities occur when the design loses both a spatial dimension and becomes a 2D structure (Fig.~S2). These boundaries are expressed in the design space as: $0<L_c<\infty$, $0<H<L_c/2$, $0<R<\infty$, and $0<\gamma<2\pi$. However, the design parameters cannot be directly measured during fabrication. Instead, we can express the boundary for non-singular mechanisms using the link lengths. A valid design is checked in two stages. First, we verify that the links $2A$, $B$, and $C$ form an obtuse triangle where $C$ is the largest side. This can be expressed as $4A^2+B^2<C^2<2A+B$. If these lengths form either a right triangle or a line, the design is singular. Then, we verify that the choice for link $D$ forms a well-defined triangular bipyramid. The lengths for $D$ are bounded between $0<D^2<2\cos^{-1}\!\left(\tfrac{4A^2+C^2-B^2}{4AC}\right)$.

\subsection*{Fabrication without external measurement}

While many FPM designs can attenuate fabrication errors, most still rely on using external measurement tools during fabrication. However, we show there is a set of normalized link lengths, $A^*= 1$, $B^* = \sqrt{5}$, $C^*=\sqrt{13}$, and $D^*=2\sqrt{2+\sqrt{2}}$, corresponding to the normalized design parameters $H^* = 1/4$, $D^* = 1/2$, $\gamma^* =\pi/2$, which can be built without measuring link lengths, using an iterative self-referencing process, and achieve a near-optimal $S_k^* = 0.073$. This process is based around a set of polygons that can only be formed with integer multiples of $A^*$ and $B^*, C^*, D^*$ (Fig. S\ref{fig:msgd}B). We use an additional intermediate link $K^*=2\sqrt{2}$ to define $D^*$. Links are copied from a static reference (Fig. S\ref{fig:msgd} A) using a tactile length comparison approach (Movie~S4). An initial seed link $A_0 = 1$ of unit length is arbitrarily chosen and copied to create $A^*,2A^*,3A^*$, using a separate iterative process (Fig.~S\ref{fig:link_scaling}). We initialize the remaining links, $B_0 = 2$, $C_0 = 3$, $H_0 = 2$, $D_0 = 3$ by rounding down the irrational values for $B^*$, $C^*$, $K^*$, and $D^*$. We refine each link type by estimating its design error $\delta$ as the distance needed to close its feedback polygon (Fig. S\ref{fig:msgd}B). Each link is updated by $\alpha$ and the refinement process repeats until the polygon can be physically closed. The process is repeated for all remaining link types using their respective polygons (Fig. S\ref{fig:msgd}C) until the design error converges to $\delta_{min}$.

To verify our results, we repeated the process using five initial seed links and averaged the results across the $n=5$ trials (Fig. S\ref{fig:msgd}C). Once each design converged, we measured $\bar{\delta_i} \%$ as the mean design error from all of the link types at the $i^{th}$ iteration step. We simulated the predicted performance $\bar{S_{k,i}}$ at each iteration step using these lengths. After only one iteration, the relative flatness $\bar{S_{k,1}} = 0.077$ converged almost entirely despite significant design error $\bar{\delta_0} = 14.7\%$ remaining. Over the next two iterations, the design continued to improve $\bar{\delta_3} = 7.82\%$, while $\bar{S_{k,3}} = 0.074$ remains close to $S_k^*$. This design stability around the point $S_k^*$ emerges from a non-linear relationship between $\delta$ and $S_k$. In this region, large errors in the design parameters have a small influence on $S_k$.

\subsection*{Integer multiple scaling of links}
We used an iterative fabrication strategy to create links of integer multiple lengths (Fig.~S\ref{fig:link_scaling}A). We begin by creating $n$ copies of the link we want to scale. To scale the $A$ link to length $nA$, we create a chain of copied $A$ links of length $n$ and apply tension to each side. Once the chain is taut, we fixture the first and last joints on a plank of wood using hot glue. We use the distance between these joints to cut an approximation of $nA_0$. We then iteratively evaluate the length and cut an updated link. For the $j$th iteration step, we estimate the design error $\delta_j$ and update the length by $\alpha_j$. We repeat the process until the link length converges such that it fits between the reference joints. Over $n=5$ trials, we found that the mean design error converged after two iteration steps with a final value of $\bar{\delta}=1.33\%$ (Fig.~S\ref{fig:link_scaling}B).

\subsection*{Review of micro-positioning structures}

We compared the compactness of the micro-FPM against other flexure positioning systems by collecting data on the footprint and workspace of recent designs Table S1. We defined the footprint as the area of the $xy$ bounding box that contained only the movement stage. For each study, we used the provided metrics for workspace. However, if the authors did not include dimensions for the footprint, we measured the values from images. To find the footprint's horizontal width $w$ and vertical height $h$, we used the author's published figures and included scale bars and converted pixel coordinates to lengths. We defined $x$ from left to right and $y$ from up to down with the origin in the top left corner of the image.

The design presented by Xi et al. Fig. 5A \cite{Xi2016-wr} had dimensions of $w = \qty{1163}{\micro\meter}$ and $h = \qty{1196}{\micro\meter}$ based a bounding box from pixels $(112,142)$ to $(326,362)$ and a conversion of \qty{5.435}{\micro\meter}/pixel. For Kim et al. we used Fig. 2A \cite{Kim2012-dp} and calculated $w = \qty{3631}{\micro\meter}$ and $h = \qty{3469}{\micro\meter}$ from pixels $(224,18)$ to $(805,573)$ with a conversion of \qty{6.25}{\micro\meter}/pixel. For Maroufi et al. we used Fig. 1 \cite{Maroufi2015-oj} and calculated $w = \qty{2412}{\micro\meter}$ and $h = \qty{2518}{\micro\meter}$ from pixels $(211,141)$ to $(416,355)$ with a conversion of \qty{11.76}{\micro\meter}/pixel. For Wan et al we used Fig. 8A \cite{Wan2016-ns} and calculated $w = \qty{245306}{\micro\meter}$ and $h = \qty{226530}{\micro\meter}$ from pixels $(296,32)$ to $(897,587)$ with a conversion of \qty{408.16}{\micro\meter}/pixel. For Li et al we used figure Fig. 8 \cite{Li2010-ml} and calculated $w = \qty{96188}{\micro\meter}$ and $h = \qty{96000}{\micro\meter}$ from pixels $(296,32)$ to $(897,587)$ with a conversion of \qty{187.50}{\micro\meter}/pixel. For Olfatnia et al we used figure Fig. 6 \cite{Olfatnia2013-iu} and calculated $w = \qty{2652}{\micro\meter}$ and $h = \qty{2613}{\micro\meter}$ from pixels $(87,20)$ to $(411,405)$ with a conversion of \qty{6.45}{\micro\meter}/pixel. The remaining authors in Table S1 included the relevant dimensions for their flexure stages.

\subsection*{Forward Kinematics}

We present a solution for the FPM's equations of motion that maps the control link's spherical angles $\theta$ and $\phi$ to the end point's Euclidean position $x$, $y$, and $z$ relative to its local coordinate frame $\boldsymbol{O}_0$. Our kinematic solution is composed of three steps (Fig.~\ref{fig:kinematics}).

\paragraph{Definitions and notation.} For the $i^{th}$ joint's local coordinate frame $\boldsymbol{O}_i$, we find the homogeneous transformation matrix $\boldsymbol{H}^0_i$ which encodes the global position of the joint. The transformation between coordinate frames $\boldsymbol{O}_i$ and $\boldsymbol{O}_j$ is defined as $\boldsymbol{H}^j_i$ and is composed of relative rotations and translations along links. We define the 3D rotation matrices for the $x$, $y$, and $z$ directions using the angles $\phi_i$, $\theta_i$, and $\psi_i$ as:

\begin{align}
\boldsymbol{R}_x(\phi_i) &= \begin{bmatrix} 1 & 0 & 0 \\ 0 & \cos(\phi_i) & -\sin(\phi_i) \\ 0 & \sin(\phi_i) & \cos(\phi_i) \end{bmatrix} \\
\boldsymbol{R}_y(\theta_i) &= \begin{bmatrix} \cos(\theta_i) & 0 & \sin(\theta_i) \\ 0 & 1 & 0 \\ -\sin(\theta_i) & 0 & \cos(\theta_i) \end{bmatrix} \\
\boldsymbol{R}_z(\psi_i) &= \begin{bmatrix} \cos(\psi_i) & -\sin(\psi_i) & 0 \\ \sin(\psi_i) & \cos(\psi_i) & 0 \\ 0 & 0 & 1 \end{bmatrix}
\end{align}

We then define the homogeneous transformation matrix between the $i^{\text{th}}$ and $j^{\text{th}}$ nodes, connected with a link of length $L_i$ along the $z$-axis, as:

\begin{align}
\boldsymbol{R}_{xyz}(\phi_i, \theta_i, \psi_i) &= \boldsymbol{R}_x(\phi_i) \cdot \boldsymbol{R}_y(\theta_i) \cdot \boldsymbol{R}_z(\psi_i) \\
\boldsymbol{T}_z(L_i) &= \begin{bmatrix} 0 \\ 0 \\ L_i \end{bmatrix} \\
\boldsymbol{H}^j_i(\phi_i, \theta_i, \psi_i, L_i) &= \begin{bmatrix} \boldsymbol{R}_{xyz}(\phi_i, \theta_i, \psi_i) & \boldsymbol{T}_z(L_i) \\ \boldsymbol{0}^\top & 1 \end{bmatrix}
\end{align}

\paragraph{Using a virtual link $L_2$ to solve for $\boldsymbol{O}_2$.} First, we find the transformation across the ground link, which is by definition $\boldsymbol{H}^1_0 = \boldsymbol{H}(0,0,0,A_0)$. Next, we introduce a virtual link with length $L_2$ between $\boldsymbol{O}_0$ and $\boldsymbol{O}_2$ and solve for $\boldsymbol{H}^2_0$ as a transformation along this link instead of along $A_1$. This approach creates three triangles between the virtual link and the $B$ and $C$ type links (Fig.~\ref{fig:kinematics}B) which simplifies future calculations for $\boldsymbol{H}^3_2$, $\boldsymbol{H}^4_2$, and $\boldsymbol{H}^5_2$. Using the laws of cosines and sines, we can solve for the complete transformation for an arbitrary choice of $\theta$ and $\phi$:

\begin{align}
L_2 &= \sqrt{A_0^2 + A_1^2 - 2A_0A_1\cos(\pi - \theta)} \\
\phi_2 &= \phi \\
\theta_2 &= \arcsin(\frac{A_1 \sin(\pi - \theta)}{L_2}) \\
\psi_2 &= 0 \\
\boldsymbol{H}^2_0 &= \boldsymbol{H}(\phi_2, \theta_2, \psi_2, L_2) \cdot \boldsymbol{R}_z(-\phi)
\end{align}

\paragraph{Solving for $\boldsymbol{O}_3$, $\boldsymbol{O}_4$, and $\boldsymbol{O}_5$.} Next, we solve for coordinate frame $\boldsymbol{O}_3$, $\boldsymbol{O}_4$, and $\boldsymbol{O}_5$ simultaneously. We begin by using the three triangles we created from the virtual link to solve for $\theta_i$ and $\psi_i$ using law of cosines.

\begin{align}
\theta_3 &= 0 \\
\psi_3 &= \pi - \arccos\left(\frac{L_2^2 + B_0^2 - C_0^2}{2L_2B_0}\right) \\
\theta_4 &= \arccos\left(\frac{L_2^2 + B_1^2 - C_1^2}{2L_2B_1}\right) - \pi \\
\psi_4 &= 0 \\
\theta_5 &= 0 \\
\psi_5 &= \arccos\left(\frac{L_2^2 + B_2^2 - C_2^2}{2L_2B_2}\right) - \frac{\pi}{2}
\end{align}

We then find $\phi_3$ and $\phi_5$ from the projections of the $B_0, B_1, B_2, D_0,$ and $D_1$ onto the $xy$ plane of the $\boldsymbol{O}_2$ coordinate frame. The projections of the $B$ links are $B'_0 = B_0 \sin(\theta_3)$, $B'_1 = B_1 \sin(\theta_4)$, $B'_2 = B_2 \sin(\theta_5)$. Without length errors in the structure, the $D$ links are parallel to the $xy$ of $O_2$. However, to account for fabrication errors in the $C$ link lengths, we must break this assumption. In this case, the projection of the links is $D'_0 = D_0 \cos(\tan(\frac{C_1 \cos(\pi-\theta_4)-C_0 \cos(\pi-\psi_3)}{D_0}))$, and $D'_1 = D_1 \cos(\tan(\frac{C_1 \cos(\pi-\theta_4)-C_2 \cos(\pi-\psi_5)}{D_1}))$. Using law of cosines, we can now solve for the rotation angles:

\begin{align}
\phi_3 &= \frac{\pi}{2} - \arccos\left(\frac{B'^{2}_0 + B'^{2}_1 - D'^{2}_0}{2B'_0 B'_1}\right) \\
\phi_4 &= 0 \\
\phi_5 &= \arccos\left(\frac{B'^{2}_2 + B'^{2}_1 - D'^{2}_1}{2B'_2 B'_1}\right) - \frac{\pi}{2}
\end{align}

The resulting transformation for $O_3, O_4,$ and $O_5$ are given by:
\begin{align}
\boldsymbol{H}^3_0 &= \boldsymbol{H}^2_0 \cdot \boldsymbol{H}(\phi_3, \theta_3, \psi_3, B_0) \\
\boldsymbol{H}^4_0 &= \boldsymbol{H}^2_0 \cdot \boldsymbol{H}(\phi_4, \theta_4, \psi_4, B_1) \\
\boldsymbol{H}^5_0 &= \boldsymbol{H}^2_0 \cdot \boldsymbol{H}(\phi_5, \theta_5, \psi_5, B_2)
\end{align}

\paragraph{Solving for $x,y,z$ using the intersection of 3 spheres.}  We now solve for the end point by finding the intersection of three spheres. While the solution has two intersection points, we know the end point must exist at the positive value.

\begin{align}
(x - x_3)^2 + (y - y_3)^2 + (z - z_3)^2 &= B_3^2 \implies \nonumber \\ x^2 + y^2 + z^2 - 2x_3x - 2y_3y - 2z_3z &= B_3^2 - x_3^2 - y_3^2 - z_3^2 \quad \label{eq:s1} \\
(x - x_4)^2 + (y - y_4)^2 + (z - z_4)^2 &= B_4^2 \implies \nonumber \\ x^2 + y^2 + z^2 - 2x_4x - 2y_4y - 2z_4z =& B_4^2 - x_4^2 - y_4^2 - z_4^2 \quad \label{eq:s2} \\
(x - x_5)^2 + (y - y_5)^2 + (z - z_5)^2 &= B_5^2 \implies \nonumber \\ x^2 + y^2 + z^2 - 2x_5x - 2y_5y - 2z_5z =& B_5^2 - x_5^2 - y_5^2 - z_5^2 \quad \label{eq:s3}
\end{align}

Let $w_i = x_i^2 + y_i^2 + z_i^2 - B_i^2$,

\begin{align}
(\ref{eq:s1})-(\ref{eq:s2}): (x_3 - x_4)x + (y_3 - y_4)y + (z_3 - z_4)z &= \frac{w_3 - w_4}{2} \label{eq:a} \\
(\ref{eq:s1})-(\ref{eq:s3}): (x_3 - x_5)x + (y_3 - y_5)y + (z_3 - z_5)z &= \frac{w_3 - w_5}{2} \label{eq:b} \\
(\ref{eq:s2})-(\ref{eq:s3}): (x_4 - x_5)x + (y_4 - y_5)y + (z_4 - z_5)z &= \frac{w_4 - w_5}{2}
\end{align}

From $\ref{eq:a} - \ref{eq:b}$:
\begin{align}
y &= a_1z + b_1 \quad \label{eq:c}\\
a_1 &= \frac{2}{d}(z_3(x_4 - x_5) + z_4(x_5 - x_3) +  z_5(x_3 - x_4)) \\
b_1 &= \frac{1}{d}(x_3(w_5 - w_4) + x_4(w_3 - w_5) + x_5(w_4 - w_3))  \\
x &= a_2z + b_2 \quad \label{eq:d} \\
a_2 &= -\frac{2}{d}(z_3(y_4 - y_5) + z_4(y_5 - y_3) + z_5(y_3 - y_4)) \\
b_2 &= -\frac{1}{d}(y_3(w_5 - w_4) + y_4(w_3 - w_5) + y_5(w_4 - w_3)) \\
d &= 2(y_3(x_4 - x_5) + y_4(x_5 - x_3) + y_5(x_3 - x_4))
\end{align}

We substitute $(\ref{eq:c})$ and $(\ref{eq:d})$ into $(\ref{eq:s1})$ and rearrange:
\begin{equation*}
(a_1^2 + a_2^2 + 1)z^2 + (2a_1(b_1 - y_3) - 2z_3 + 2a_2(b_2 - x_3))z + (z_3^2 + (b_1 - y_3)^2 + (b_2 - x_3)^2 - B_3^2) = 0
\end{equation*}

Finally, we can solve for the positive root of this quadratic equation to find $z$ and substitute this value into $(\ref{eq:c})$ and $(\ref{eq:d})$ to obtain $y$ and $x$, giving us the endpoint's position.

\subsection*{Inverse Kinematics}

Here, we derive the actuated FPM's (Fig.~\ref{fig:actuation}A) inverse kinematics, or the mapping between the motor angles $\alpha$ and $\beta$ and the end point's position $(x,y)$ in the FPM's local coordinate frame (Fig.~S\ref{fig:ik}). For the robotic FPM, the vertical motion is realized using an independent linear axis, making the solution for $z$ trivial. Our solution assumes an idealized FPM geometry with link length parameters $A,B,C,$ and $D$.

\paragraph{Definitions and intermediate angles.} For a given Cartesian target $(x,y,L_c)$ in $\boldsymbol{O}_0$, we first define the vector $\boldsymbol{X}$ and the distance $\mathbf{O D}$  (Fig.~S\ref{fig:ik}A) as:

\begin{equation}
\boldsymbol{X}
=
\begin{bmatrix}
x \\[2pt] y \\[2pt] L_c
\end{bmatrix},
\qquad
\mathbf{O D} = \|\boldsymbol{X}\|
= \sqrt{x^2 + y^2 + L_c^2}.
\end{equation}

We then parameterize $\boldsymbol{X}$ using azimuth, elevation, and polar angles (Fig.~\ref{fig:ik}A,B,C). The azimuth $\mathrm{Az}$ in the $x$–$z$ plane and the elevation $\mathrm{El}$ relative to the $x$–$z$ plane are:
\begin{align}
\mathrm{Az} &= \operatorname{atan2}\!\bigl(x,\,L_c\bigr), \\[4pt]
\mathrm{El} &= \operatorname{atan2}\!\bigl(
  y,\,
  \sqrt{x^2 + L_c^2}
\bigr).
\end{align}
The polar angle $\theta$ between $\boldsymbol{X}$ and the FPM's local $+z$ axis is
\begin{equation}
\theta
=
\operatorname{atan2}\!\left( \sqrt{x^2 + y^2},\, L_c \right).
\end{equation}

\paragraph{Solving for point $\mathbf{B}$.}
Next, we solve for the position of $\mathbf{B}$. To begin, we calculate the distance $\mathbf{O B}$, using the law of sines:
\begin{equation}
\mathbf{O B}^2
=
2A^2 - 2A^2 \cos\!\bigl(\pi - 2\theta\bigr),
\qquad
\mathbf{O B} = 2A \cos\theta
\end{equation}

We then place $\boldsymbol{X}_b$ on the sphere of radius $\mathbf{O B}$ with the same azimuth and elevation as $\boldsymbol{X}$:
\begin{align}
x_b &= \mathbf{O B} \cos(\mathrm{El}) \sin(\mathrm{Az}),\\[2pt]
y_b &= \mathbf{O B} \sin(\mathrm{El}),\\[2pt]
z_b &= \mathbf{O B} \cos(\mathrm{El}) \cos(\mathrm{Az}),
\end{align}
so that
\begin{equation}
\mathbf{B}
=
\begin{bmatrix}
x_b \\[2pt] y_b \\[2pt] z_b
\end{bmatrix},
\qquad
\|\mathbf{B}\| = \mathbf{O B}.
\end{equation}

\paragraph{Solving for the first motor angle $\alpha$.}
The first joint axis is aligned with the global $y$ axis, and is offset by link length $A$ along $+z$. We can now compute $\alpha$ from the projection of $\mathbf{B}$ onto the $x$–$z$ plane, relative to the offset $A$:
\begin{equation}
\alpha
=
\operatorname{atan2}\!\bigl(x_b,\; z_b - A\bigr).
\end{equation}

\paragraph{Solving for $\boldsymbol{O}_0$ from the intersection of a circle and plane.}

To compute the second joint angle $\beta$, we first find the intersection of plane $P$ at point $\mathbf{G}$ (Fig.~S\ref{fig:ik}B) and the circle $\mathcal{C}$ (Fig.~\ref{fig:ik}D).

We reuse the rotation matrices defined from the forward kinematics and consider a pure rotation about the $y$ axis by $\alpha$:
\begin{equation}
\boldsymbol{R}_y(\alpha)
=
\begin{bmatrix}
\cos\alpha & 0 & \sin\alpha \\
0          & 1 & 0          \\
-\sin\alpha & 0 & \cos\alpha
\end{bmatrix}.
\end{equation}
Let $\boldsymbol{e}_x = [1,\,0,\,0]^\top$ denote the unit vector along the global $x$ axis. We define the normal vector for the plane as
\begin{equation}
\boldsymbol{N}_b
=
\boldsymbol{R}_y(\alpha)\,\boldsymbol{e}_x.
\end{equation}

The circle $\mathcal{C}$ for the second joint has center $\mathbf{B}$, radius $\mathbf{OB}$, and lies in the plane orthogonal to $\boldsymbol{N}_b$:
\begin{equation}
P_\mathcal{C}
=
\left\{
  \boldsymbol{x} \in \mathbb{R}^3
  \,\middle|\,
  \|\boldsymbol{x} - \boldsymbol{X}_b\| = \mathbf{OB},\;
  \boldsymbol{N}_b^\top(\boldsymbol{x} - \boldsymbol{X}_b) = 0
\right\}.
\end{equation}

We define the unit direction of $\boldsymbol{X}$ as
\begin{equation}
\boldsymbol{N}_g
=
\frac{\boldsymbol{X}}{\|\boldsymbol{X}\|}
=
\frac{1}{\mathbf{O D}}
\begin{bmatrix}
x \\[2pt] y \\[2pt] L_c
\end{bmatrix}.
\end{equation}
We then construct a point $\mathbf{G}$ at the mid point of $\overline{\mathbf{B D}}$:
\begin{align}
\lambda
&=
\frac{\mathbf{O B} + (\mathbf{O D} - \mathbf{O B})/2}{\mathbf{O D}}
=
\frac{\mathbf{O B} + \mathbf{O D}}{2\mathbf{O D}}, \\[4pt]
\boldsymbol{X}_g
&=
\lambda\,\boldsymbol{X}.
\end{align}
We then define the plane $P$, centered at $\mathbf{G}$ and normal to $\boldsymbol{X}$ as:
\begin{equation}
P
=
\left\{
  \boldsymbol{x} \in \mathbb{R}^3
  \,\middle|\,
  \boldsymbol{N}_g^\top(\boldsymbol{x} - \mathbf{B}) = 0
\right\}.
\end{equation}

The circle's plane $P_\mathcal{C}$ passing through $\mathbf{B}$ has equation
\begin{equation}
\boldsymbol{N}_b^\top\boldsymbol{x} = d_b,
\qquad
d_b = \boldsymbol{N}_b^\top \mathbf{B},
\end{equation}
and the plane $P$ has equation
\begin{equation}
\boldsymbol{N}_g^\top\boldsymbol{x} = d_g,
\qquad
d_g = \boldsymbol{N}_g^\top \mathbf{G}.
\end{equation}
The direction vector of the line of intersection of $P_\mathcal{C}$ and $P$ is
\begin{equation}
\boldsymbol{d}
=
\boldsymbol{N}_b \times \boldsymbol{N}_g.
\end{equation}
If $\|\boldsymbol{d}\| = 0$, the planes are parallel and the intersection is either empty or infinite. Otherwise, we construct a particular point $\boldsymbol{p}_0$ on the intersection line as
\begin{equation}
\boldsymbol{p}_0
=
\frac{
  d_b\left(\boldsymbol{N}_g \times \boldsymbol{d}\right)
  +
  d_g\left(\boldsymbol{d} \times \boldsymbol{N}_b\right)
}{
  \|\boldsymbol{d}\|^2
}.
\end{equation}
All points on the intersection line can then be written as
\begin{equation}
\boldsymbol{x}(t)
=
\boldsymbol{p}_0 + t\,\boldsymbol{d},
\qquad
t \in \mathbb{R}.
\end{equation}

To enforce the circle constraint, we require
\begin{equation}
\|\boldsymbol{x}(t) - \mathbf{B}\|^2 = B^2.
\end{equation}
Let
\begin{equation}
\boldsymbol{p}_0'
=
\boldsymbol{p}_0 - \mathbf{B}.
\end{equation}
Then
\begin{align}
\|\boldsymbol{x}(t) - \mathbf{B}\|^2
&=
\|\boldsymbol{p}_0' + t\,\boldsymbol{d}\|^2 \\[2pt]
&=
(\boldsymbol{d}^\top\boldsymbol{d})\,t^2
+
2(\boldsymbol{p}_0'^\top\boldsymbol{d})\,t
+
\boldsymbol{p}_0'^\top\boldsymbol{p}_0'.
\end{align}
Imposing $\|\boldsymbol{x}(t) - \mathbf{B}\|^2 = B^2$ gives the scalar quadratic
\begin{equation}
a t^2 + b t + c = 0,
\end{equation}
with
\begin{align}
a &= \boldsymbol{d}^\top\boldsymbol{d}, \\[2pt]
b &= 2\,\boldsymbol{p}_0'^\top\boldsymbol{d}, \\[2pt]
c &= \boldsymbol{p}_0'^\top\boldsymbol{p}_0' - B^2.
\end{align}
The discriminant is
\begin{equation}
\Delta = b^2 - 4ac.
\end{equation}
If $\Delta < 0$ there is no real intersection. For $\Delta \ge 0$, the two solutions are
\begin{equation}
t_{1,2}
=
\frac{-b \pm \sqrt{\Delta}}{2a}.
\end{equation}
The corresponding intersection points on the circle are
\begin{equation}
\boldsymbol{X}_1 = \boldsymbol{p}_0 + t_1 \boldsymbol{d},
\qquad
\boldsymbol{X}_2 = \boldsymbol{p}_0 + t_2 \boldsymbol{d}.
\end{equation}

By default, we use the positive root of the quadratic, which corresponds to point $\mathbf{A}$, unless the circle is tangent to the plane:
\begin{equation*}
\mathbf{A}
=
\begin{cases}
\boldsymbol{X}_2,& \text{if two distinct solutions exist},\\[2pt]
\boldsymbol{X}_1,& \text{if the circle is tangent and there is a single solution}.
\end{cases}
\end{equation*}

We define the components of $\mathbf{A}$ as:

\begin{equation}
\mathbf{A} = \begin{bmatrix}
x_\mathbf{A} \\[2pt] y_\mathbf{A} \\[2pt] z_\mathbf{A}
\end{bmatrix}
\end{equation}

\paragraph{Solving for the second motor angle \texorpdfstring{$\beta$}{beta}.}

Finally, we solve for $\beta$ using the projection of $\mathbf{A}$ onto the $y$–$z$ plane:
\begin{equation}
\beta
=
\operatorname{atan2}\!\bigl(y_\mathbf{A},\,z_\mathbf{A}\bigr).
\end{equation}

\subsection*{Parallelism with the Z-axis}

We evaluated the robotic FPM’s Z-axis parallelism by probing XY planes at \qty{25}{\mm} increments over a \qty{200}{\mm} vertical range relative to a granite reference surface using gauge blocks (Fig.~S\ref{fig:parallel}). Each plane was fit from a $3\times3$ grid of points spanning a $\qty{100}{\mm} \times \qty{100}{\mm}$ workspace. The tilt of each plane’s normal vector $\boldsymbol{n}_i$ relative to the granite surface normal $\boldsymbol{n}_g$ was computed as
\begin{equation}
\theta_{\mathrm{tilt}} = \cos^{-1}\!\left(
\frac{\boldsymbol{n}_i \cdot \boldsymbol{n}_g}
{\|\boldsymbol{n}_i\|\,\|\boldsymbol{n}_g\|}
\right).
\end{equation}
Across the full Z range, the maximum measured tilt was $0.021^\circ$, corresponding to a lateral runout of $\pm\qty{18.3}{\um}$ over a $\pm\qty{50}{\mm}$ translation. This indicates that planar motion maintains tight parallelism throughout the Z-axis travel.

\setcounter{figure}{0} 

\section*{Supplemental Figures}
\begin{figure}[H]
  \centering
  \makebox[\textwidth][c]{\includegraphics[width=\textwidth]{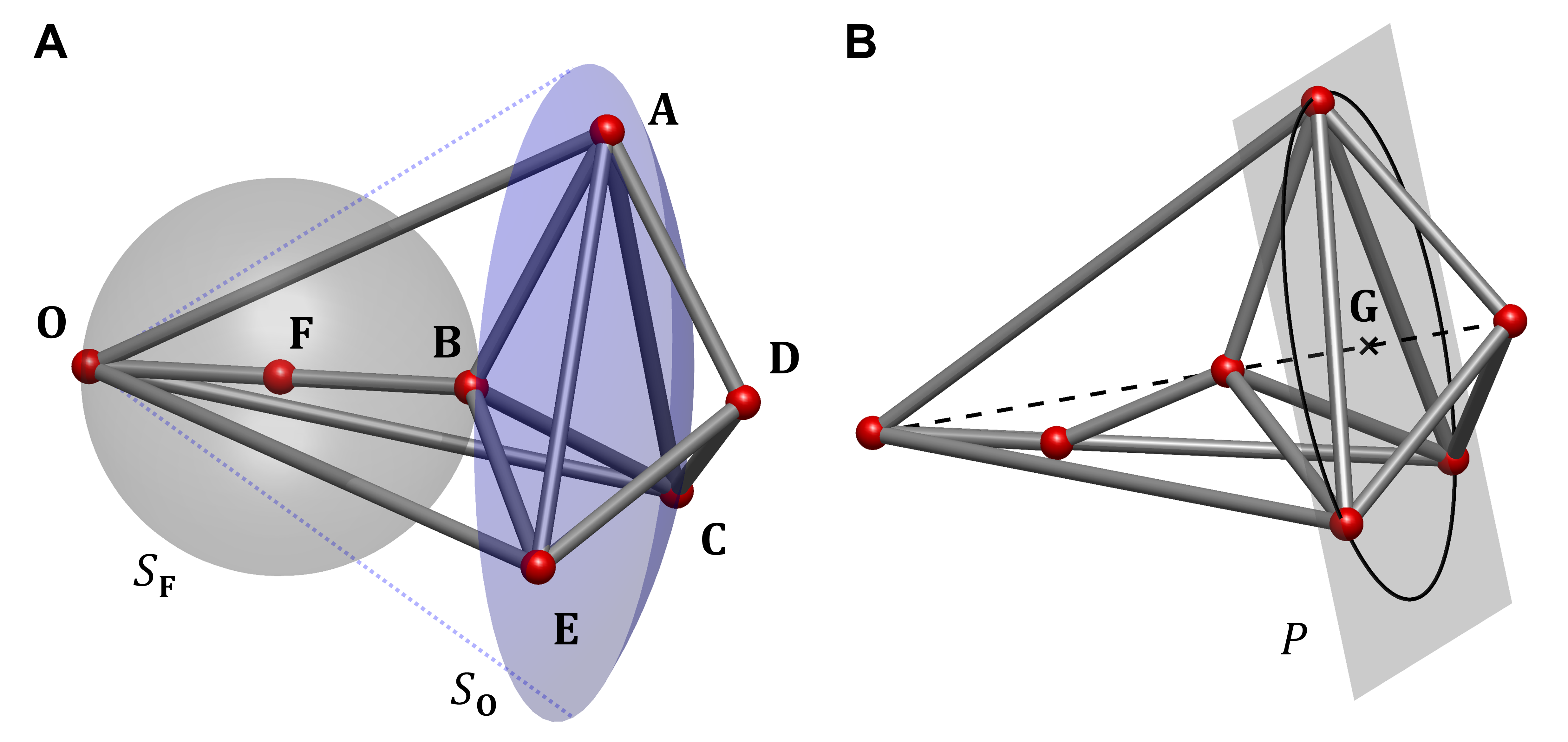}}
  \caption{Fig. S1. \textbf{FPM proof.} The geometry of the FPM with labeled points and reference surfaces used in the proof (\textbf{A}-\textbf{B}).}
  \label{fig:proof}
\end{figure}

\begin{figure}[H]
  \centering
    \makebox[\textwidth][c]{\includegraphics[width=\textwidth]{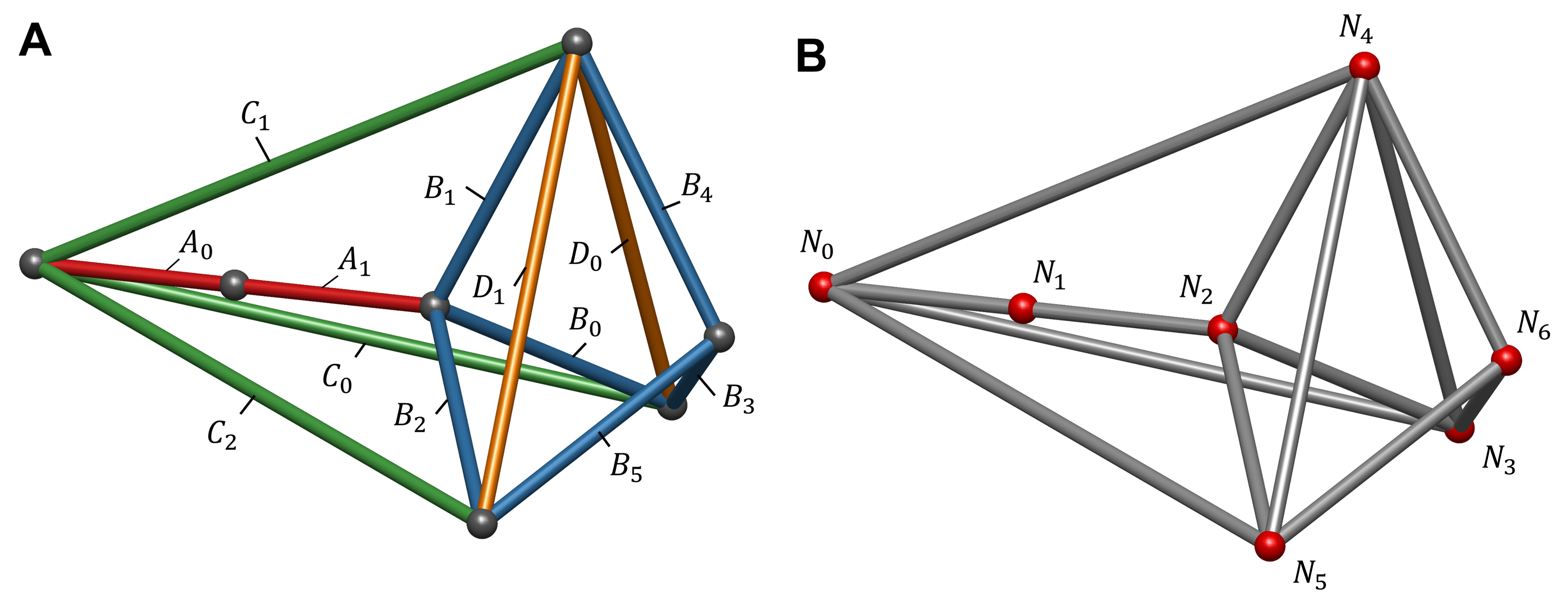}}
  \caption{Fig. S2. \textbf{FPM naming conventions.} The FPM naming convention for \textbf{(A)} links and \textbf{(B)} nodes/joints.}
  \label{fig:naming}
\end{figure}

\begin{figure}[H]
  \centering
    \makebox[\textwidth][c]{\includegraphics[width=\textwidth]{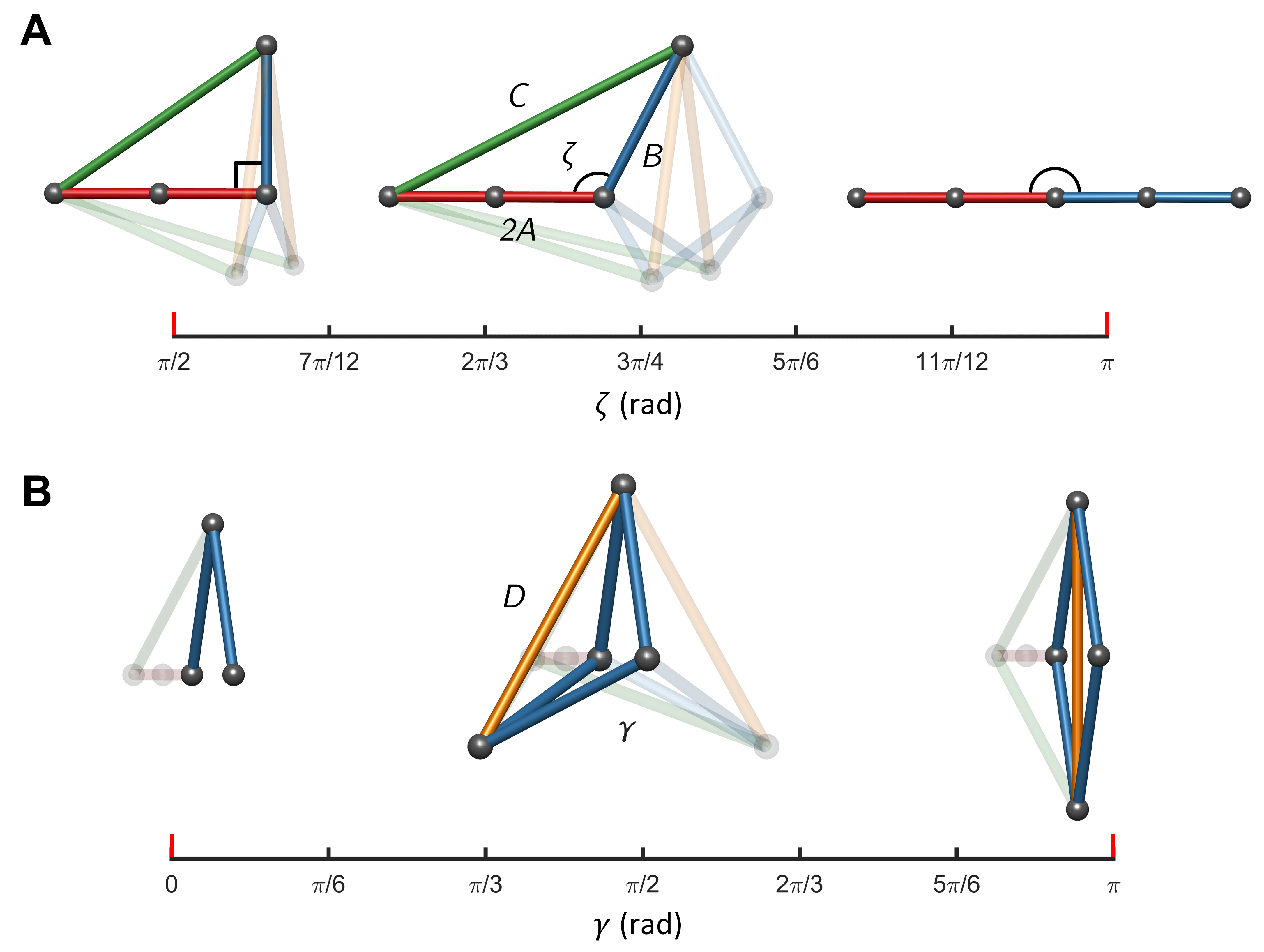}}
  \caption{Fig. S3. \textbf{Design space limits.} The valid design space for the FPM is defined between four singular design configurations. A singular design occurs when \textbf{(A)} the triangle $\triangle2ABC$ collapses into a line or \textbf{(B)} $\gamma={0,\pi}$.}
  \label{fig:limits}
\end{figure}

\begin{figure}
\makebox[\textwidth][c]{\includegraphics[width=\textwidth]{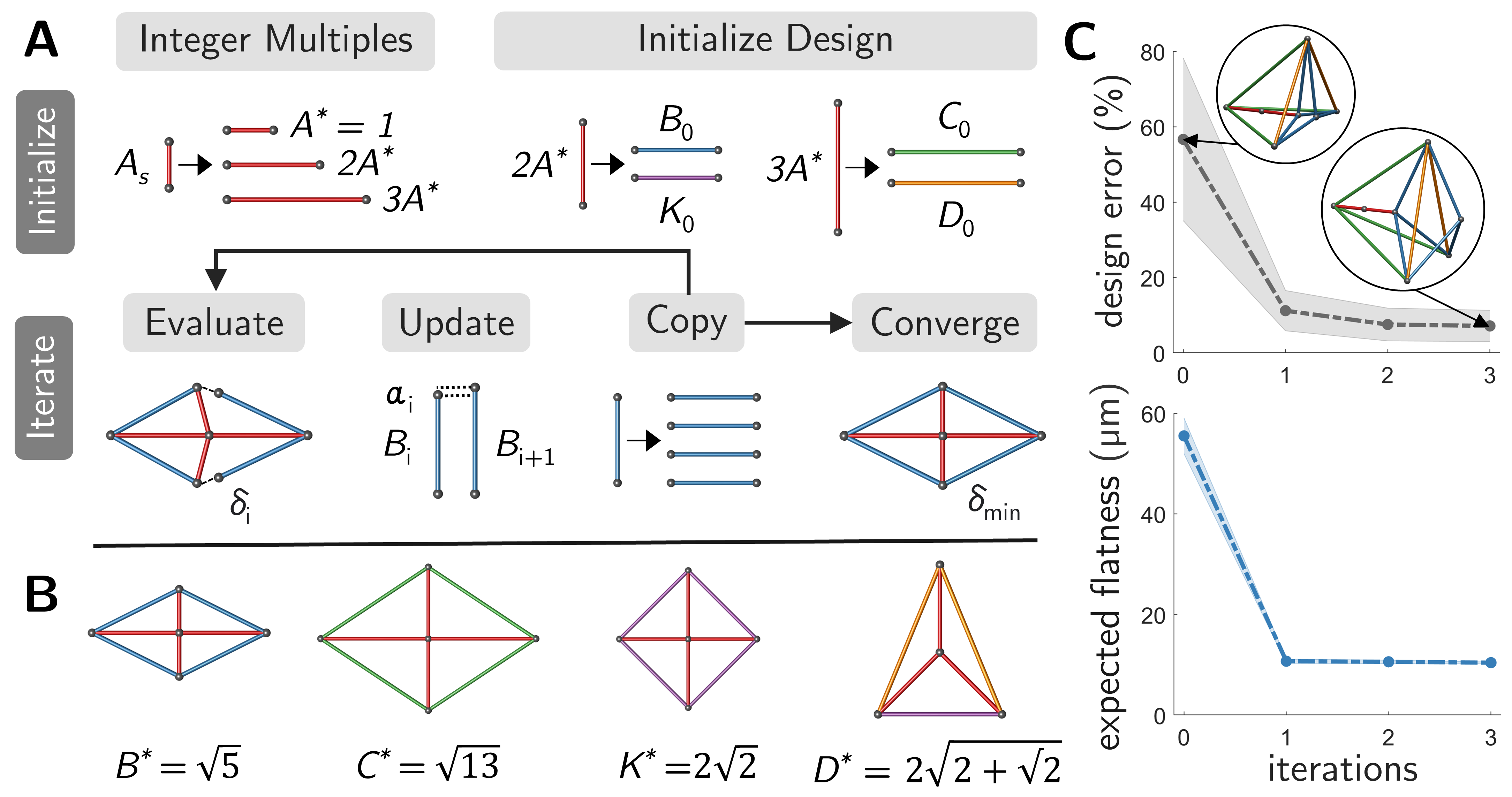}}
\caption{Fig.~S4. \textbf{Fabrication without external metrology.}
{\bf (A)} Starting from an arbitrarily chosen seed link $A_s$, integer copies of $A^*$ are formed and used to initialize the remaining links. Each link length is iteratively refined toward its target value using feedback polygons that only close at the optimal geometry. At iteration $i$, the design error $\delta_i$ is estimated from the polygon closure error and used to update the link by $\alpha_i$.
{\bf (B)} All links are updated in parallel using their respective feedback polygons; an intermediate link $K$ enables construction of the polygon defining $D$.
{\bf (C)} Experimental convergence from a poorly performing initial design to a near-optimal FPM without external measurement. The predicted relative flatness $\bar{S}_k$ is computed from the measured link lengths at each iteration. Error bars represent min and max values from $n=5$ trials.}

\label{fig:msgd}
\end{figure}

\begin{figure}[H]
  \centering
    \makebox[\textwidth][c]{\includegraphics[width=\textwidth]{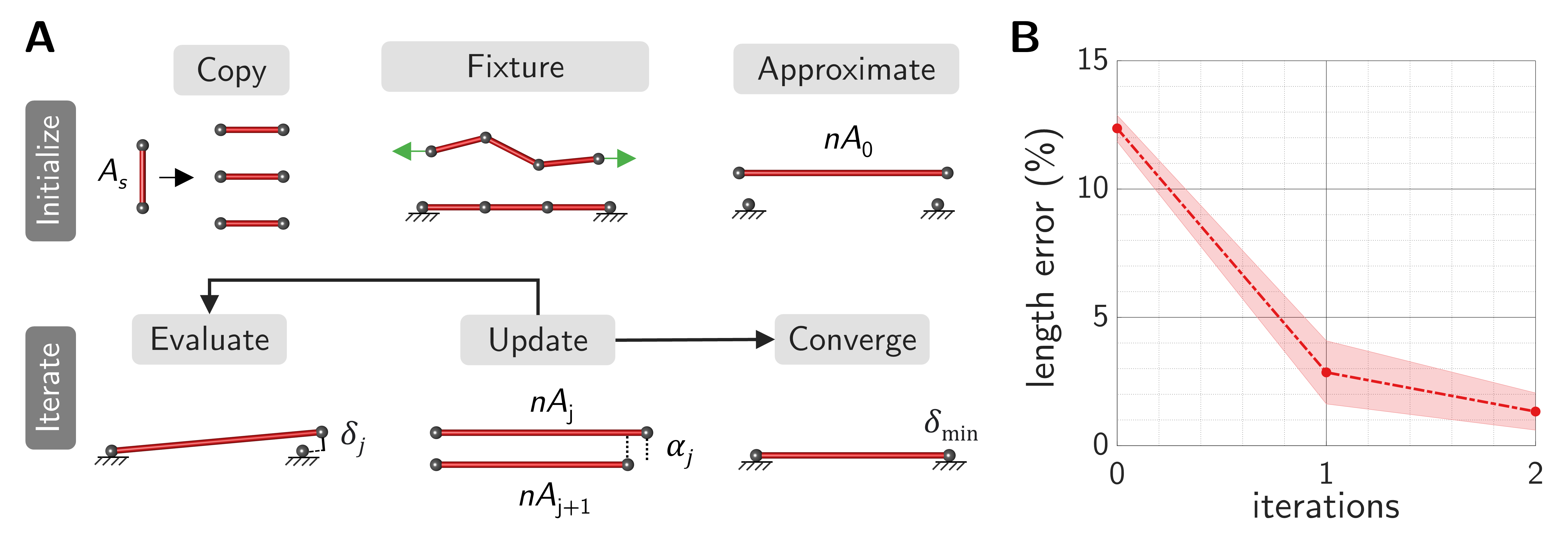}}
  \caption{Fig. S5. \textbf{Link scaling of integer lengths.} \textbf{(A)} An iterative approach to create links with lengths that are $n$ multiples of the $A$ link. \textbf{(B)} We created $2A$ and $3A$ links using this process and found the lengths converged after two iterations. Error bars represent min and max values from $n=5$ trials.}
  \label{fig:link_scaling}
\end{figure}

\begin{figure}[H]
  \centering
    \makebox[\textwidth][c]{\includegraphics[width=\textwidth]{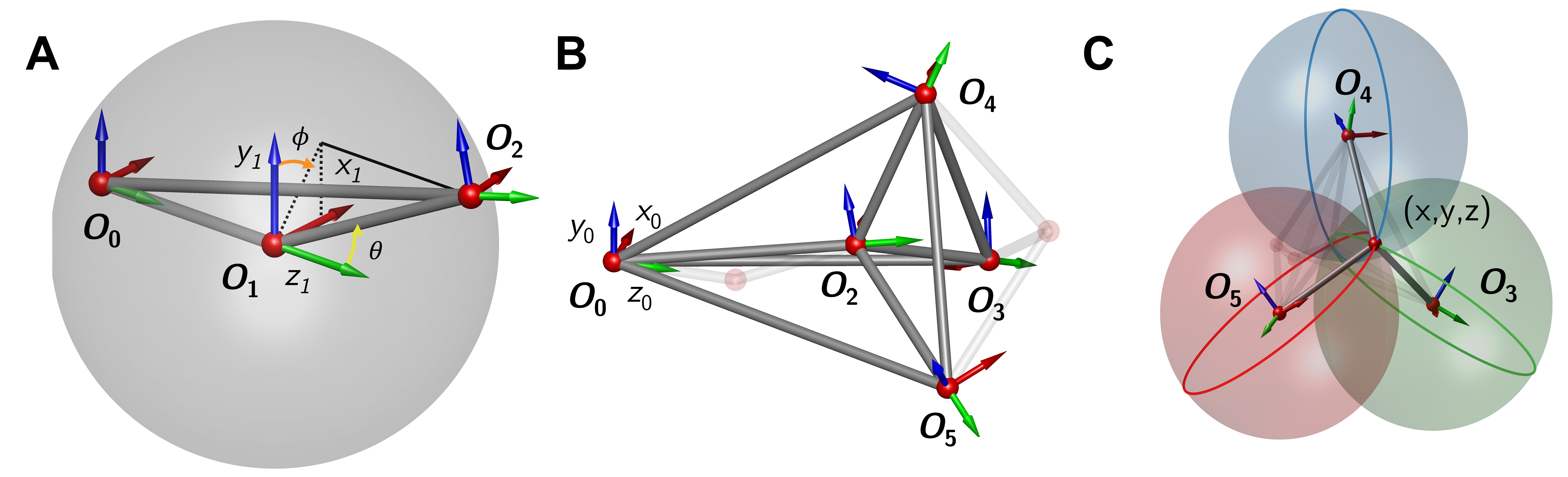}}
  \caption{Fig. S6. \textbf{Forward kinematics.} \textbf{(A)} Construct the virtual link with length $L_2$ between the ground and control links and solve for frames $O_1$ and $O_2$. \textbf{(B)} Solve for frames $O_3$--$O_5$ using triangles created by the virtual and physical links. \textbf{(C)} Find the endpoint by solving for the intersection of three spheres.}
  \label{fig:kinematics}
\end{figure}

\begin{figure}[H]
  \centering
    \makebox[\textwidth][c]{\includegraphics[width=\textwidth]{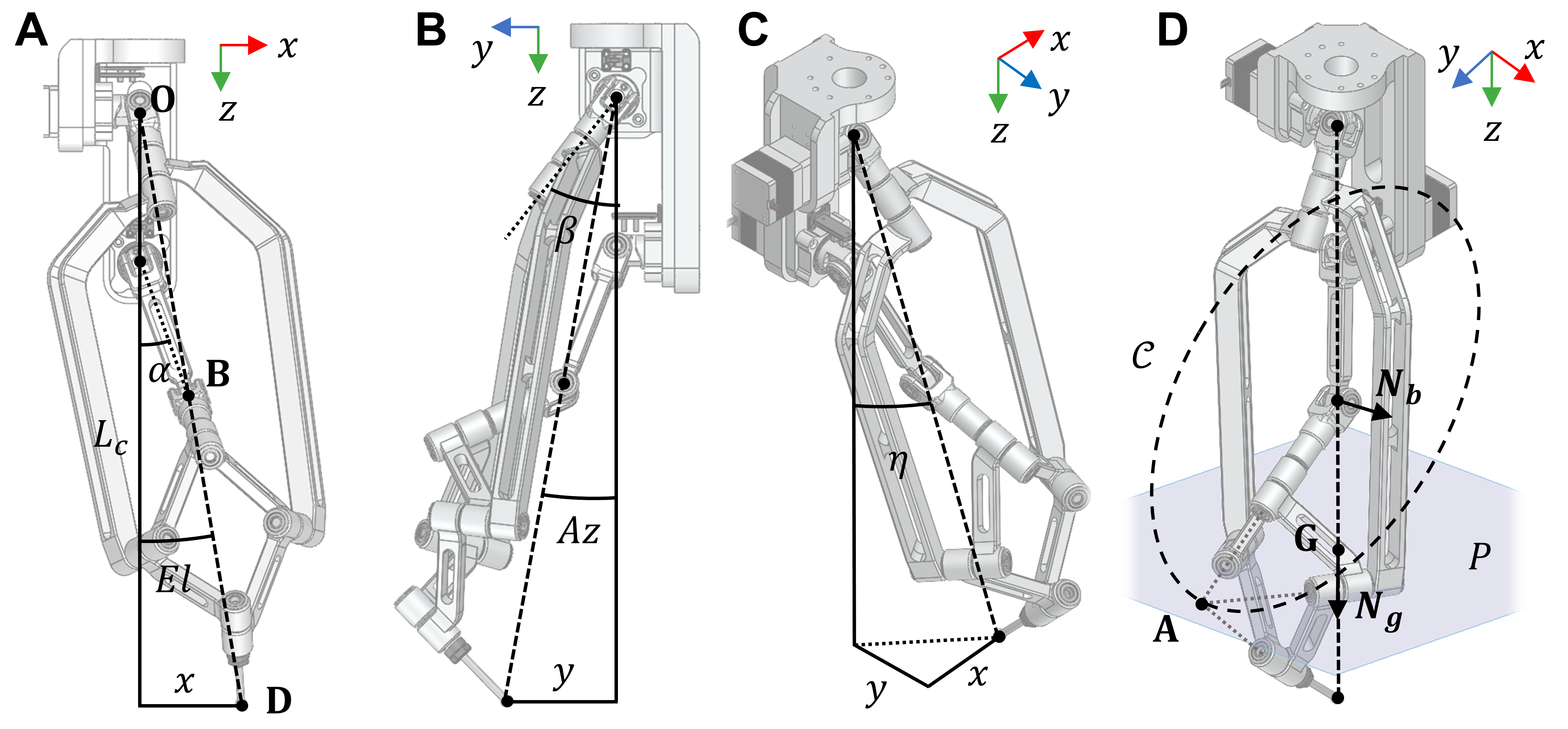}}
  \caption{Fig. S7. \textbf{Inverse kinematics.} Labeled diagrams for the FPM's inverse kinematics (\textbf{A}-\textbf{D}).}
  \label{fig:ik}
\end{figure}

\begin{figure}[H]
  \centering
    \makebox[\textwidth][c]{\includegraphics[width=0.8\textwidth]{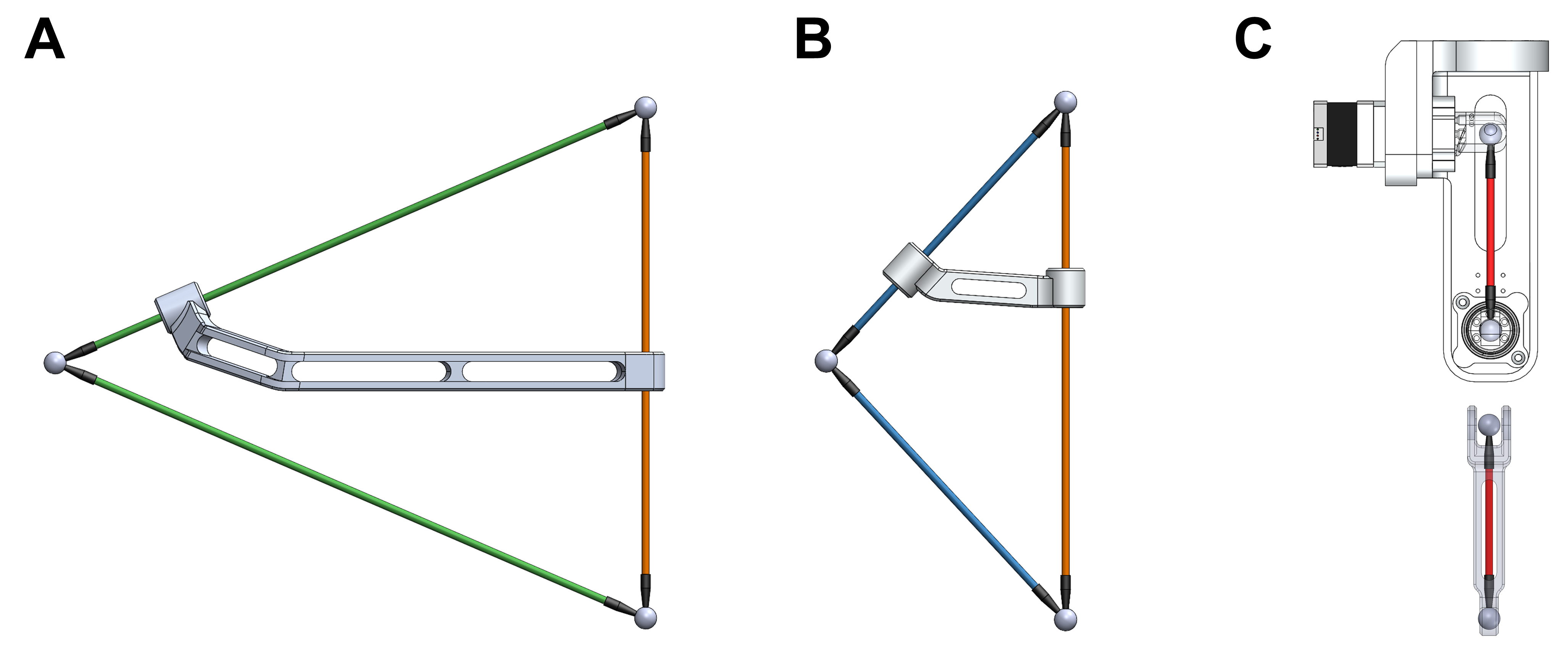}}
  \caption{Fig. S8. \textbf{Triangular decomposition of the FPM.} The robotic FPM is composed of links that enforce angle constraints between \textbf{(A)} the $C$ and $D$ links and \textbf{(B)} the $B$ and $D$ links. \textbf{(C)} The ground and control link maintain a fixed distance corresponding to the $A$ link.}
  \label{fig:triangles}
\end{figure}

\begin{figure}[H]
  \centering
    \makebox[\textwidth][c]{\includegraphics[width=0.9\textwidth]{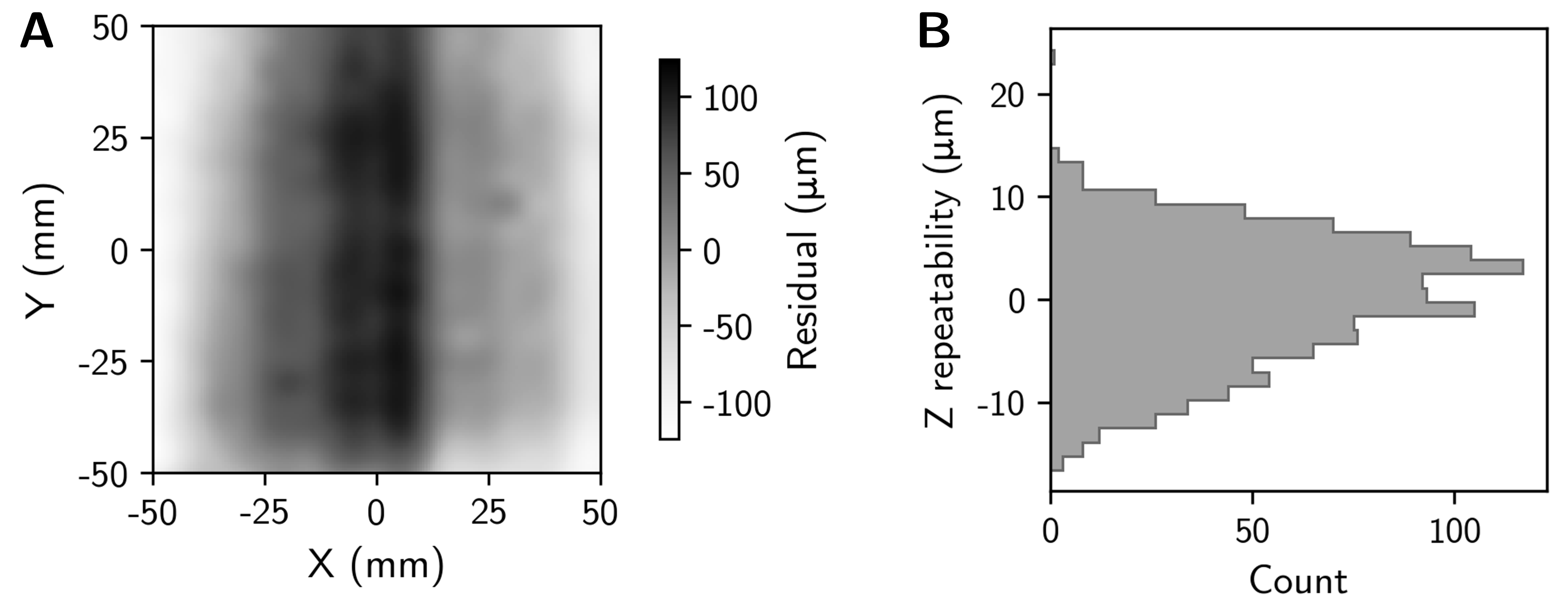}}
  \caption{Fig. S9. \textbf{Granite surface scan and repeatability.} (\textbf{A}) The surface scan of a granite slab used to compensate for Z bias. (\textbf{B}) The FPM's z probing repeatability over the $\qty{100}{\mm} \times \qty{100}{\mm}$ workspace, across a $9\times9$ grid of points scanned over $n=5$ trials.}
  \label{fig:granite}
\end{figure}

\begin{figure}[H]
  \centering
    \makebox[\textwidth][c]{\includegraphics[width=0.8\textwidth]{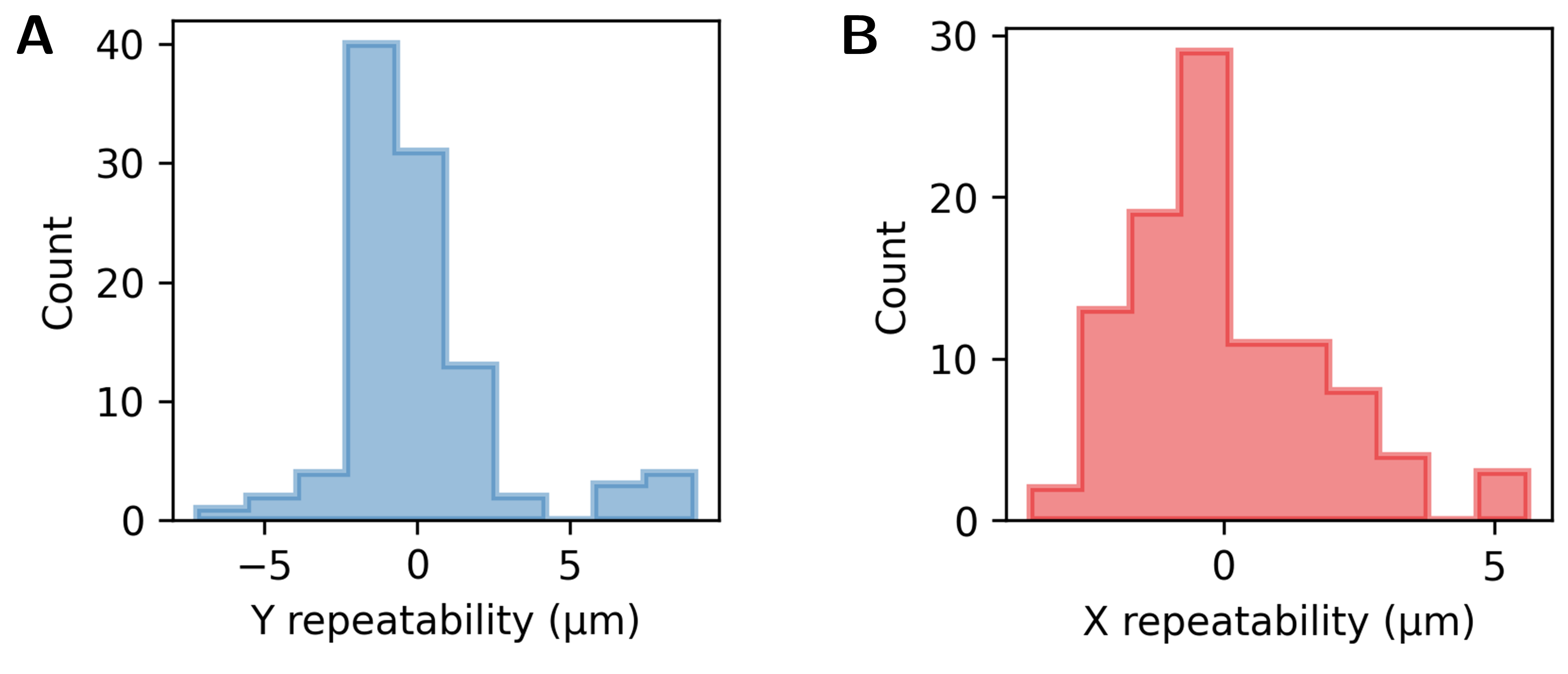}}
  \caption{Fig. S10. \textbf{Repeatability in $x$ and $y$.} The FPM's repeatability, at the home position, in the (\textbf{A}) $x$ and (\textbf{B}) $y$ directions. Histograms represent the deviations from the mean position across $n=100$ samples.}
  \label{fig:xy_repeatability}
\end{figure}


\begin{figure}[H]
  \centering
    \makebox[\textwidth][c]{\includegraphics[width=0.8\textwidth]{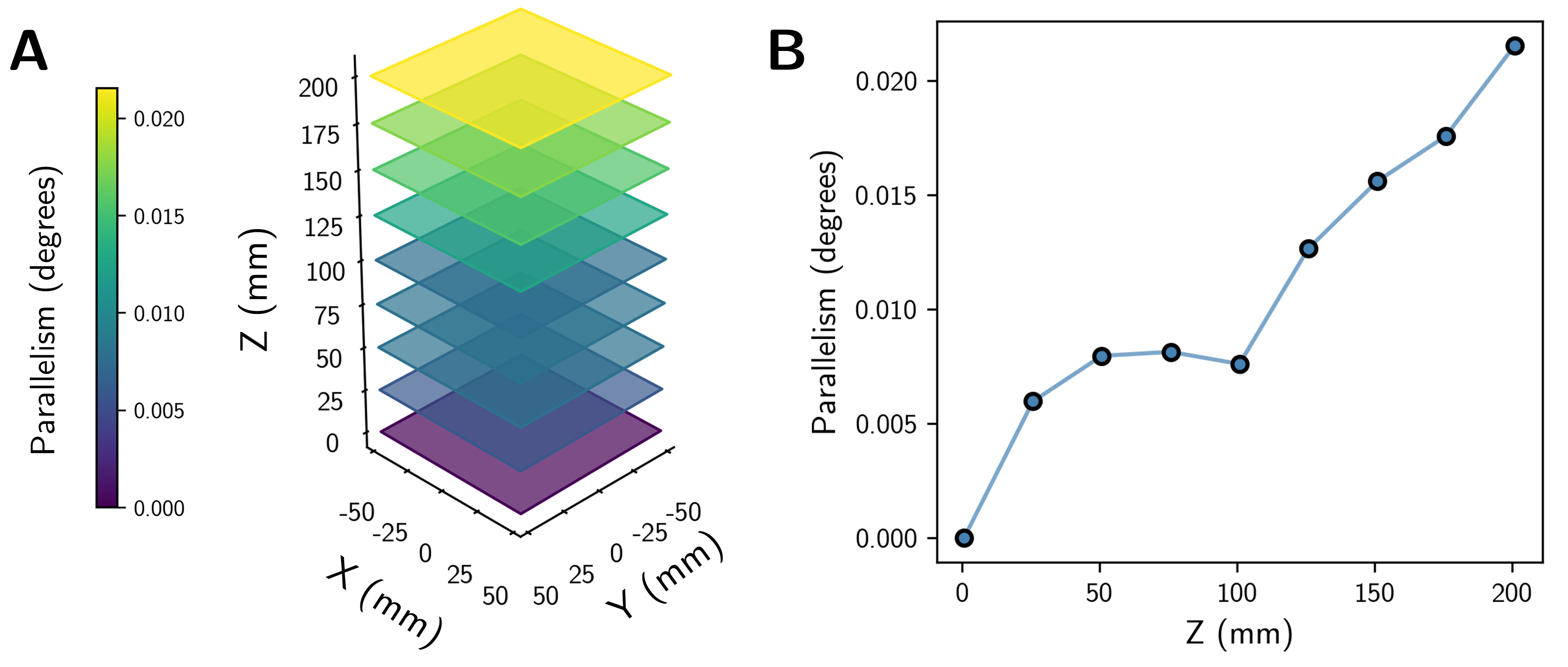}}
  \caption{Fig. S11. \textbf{Error in the z-axis.} \textbf{(A)} We quantified the parallelism error introduced by the z-axis for vertical movements. \textbf{(B)} A line plot showing the gradual tilt over a \qty{200}{\mm} range of motion.}
  \label{fig:parallel}
\end{figure}

\newpage

\section*{Supplemental Tables}
\begin{table}[H]
  \centering
  \caption{\textbf{Table S1.} Comparison of footprint ($F$) to XY workspace ($W$) ratios for flexure-based positioning systems.}
  \vspace{6pt}
  \begin{tabular}{lccc}
    \toprule
    Reference & Footprint ($\qty{}{\um}^2$) & XY Workspace ($\qty{}{\um}^2$) & Area ratio (\%) \\
    \midrule
    \cite{Xi2016-wr}       & $1.391\times10^{6}$  & $2.812\times10^{2}$  & 0.02021 \\
    \cite{Kim2012-dp}      & $1.260\times10^{7}$  & $2.500\times10^{3}$  & 0.01984 \\
    \cite{Maroufi2015-oj}  & $6.073\times10^{6}$  & $6.400\times10^{1}$  & 0.001054 \\
    \cite{Olfatnia2013-iu} & $6.929\times10^{6}$  & $3.240\times10^{4}$  & 0.4677 \\
    \cite{Li2010-ml}       & $9.230\times10^{9}$  & $1.632\times10^{4}$  & 0.0001768 \\
    \cite{Awtar2013-gs}    & $6.503\times10^{10}$ & $1.000\times10^{8}$  & 0.1538 \\
    \cite{Lyu2023-ft}      & $5.382\times10^{10}$ & $2.234\times10^{8}$  & 0.4151 \\
    \cite{Wan2016-ns}      & $5.560\times10^{10}$ & $1.960\times10^{8}$  & 0.3525 \\
    \cite{Graser2021-dd}   & $5.715\times10^{10}$ & $4.000\times10^{8}$  & 0.7001 \\
    \cite{Man2025-lu}                & $8.040\times10^{6}$  & $3.000\times10^{4}$  & 0.3731 \\
    \cite{Leveziel2022-ex}   & $4.180\times10^{8}$ & $2.500\times10^{5}$  & 0.0598 \\ 
    (This work)            & $5.281\times10^{3}$  & $1.256\times10^{3}$  & 23.79 \\

    \bottomrule
  \end{tabular}
  \label{tab:S1}
\end{table}

\begin{table}[H]
  \centering
  \caption{\textbf{Table S2.} The average link lengths and corresponding design parameters for each FPM.}
  \vspace{6pt}
  \begin{tabular}{lcccccccc}
    \toprule
    Structure & $A$ & $B$ & $C$ & $D$ & $L_c$ & $R$ & $H$ & $\gamma$ (deg) \\
    \midrule
    Micro ($\qty{}{\um}$) & 31.71 & 45.98 & 91.95 & 72.93 & 100 & 42.18 & 18.29 & 120.71 \\
    Desktop-1 (cm) & 5.21 & 11.58 & 18.69 & 18.98 & 20.67 & 10.39 & 5.13 & 96.00 \\
    Desktop-1.25 (cm) & 6.52 & 14.75 & 23.51 & 24.17 & 25.73 & 13.31 & 6.35 & 99.09 \\
    Desktop-1.75 (cm) & 8.70 & 19.22 & 31.58 & 30.42 & 36.08 & 16.80 & 9.34 & 100.55 \\
    Meter (m) & 0.68 & 1.00 & 1.97 & 1.60 & 2.12 & 0.93 & 0.38 & 123.24 \\
    Robotic FPM (mm) & 125.00 & 224.06 & 414.82 & 329.10 & 487.50 & 118.75 & 190.00 & 120.00 \\
    \bottomrule
  \end{tabular}
  \label{tab:S2}
\end{table}

\newpage

\begin{longtable}{c >{\raggedright\arraybackslash}p{0.50\textwidth} l c r}
\caption{\textbf{Table S3.} Bill of materials for the robotic FPM.}
\label{tab:bom}\\
\toprule
Item & Part Name / File Name & Vendor & Quantity & Cost (USD) \\
\midrule
\endfirsthead

\toprule
Item & Part Name / File Name & Vendor & Quantity & Cost (USD) \\
\midrule
\endhead

\midrule
\multicolumn{5}{r}{\emph{Continued on next page}} \\
\midrule
\endfoot

\bottomrule
\endlastfoot

1  & BD Link (Press Fit) & Xometry & 2 & \$345.48 \\
2  & CD Link (Right) & Xometry & 1 & \$654.72 \\
3  & BD Link & Xometry & 1 & \$269.41 \\
4  & CD Link (Left) & Xometry & 1 & \$646.30 \\
5  & BD Link (COLLET) & Xometry & 1 & \$365.59 \\
6  & Y Connector & Xometry & 2 & \$275.96 \\
7  & Drive Shaft & Xometry & 2 & \$325.86 \\
8  & A Link & Xometry & 1 & \$251.35 \\
9  & Ground Link & Xometry & 1 & \$528.56 \\
10 & Linear Rail Bracket & Xometry & 1 & \$76.60 \\
11 & Beam Bracket & Xometry & 1 & \$229.75 \\
12 & Bearing Support Block & Xometry & 2 & \$195.48 \\
13 & Aluminum Standoff & Xometry & 1 & \$127.81 \\
14 & Alpha Flag & Protolabs & 1 & \$68.07 \\
15 & Beta Flag & Protolabs & 1 & \$42.58 \\
16 & Endstop Cover & Protolabs & 2 & \$69.62 \\
17 & End Cap & Protolabs & 2 & \$141.21 \\
18 & Z-axis Frame & Vention & 1 & \$825.23 \\
19 & Linear Z-axis Rail & \href{https://a.co/d/bD13QJD}{Amazon} & 1 & \$330.00 \\
20 & 50:1 Harmonic Gearbox & \href{https://a.co/d/gZQekD7}{Amazon} & 2 & \$199.98 \\
21 & ER11 8mm Collet & \href{https://a.co/d/cKEqJ7A}{Amazon} & 1 & \$11.45 \\
22 & BIGTREETECH Octopus V1.1 & \href{https://a.co/d/bDXvKU8}{Amazon} & 1 & \$199.98 \\
23 & BIGTREETECH TMC5160T & \href{https://a.co/d/gO8jZhW}{Amazon} & 1 & \$49.99 \\
24 & STEPPERONLINE 0.9deg Nema 17 & \href{https://a.co/d/d5oiLf1}{Amazon} & 2 & \$33.98 \\
25 & Raspberry Pi 3B+ & \href{https://a.co/d/cPRH2yU}{Amazon} & 1 & \$54.00 \\
26 & PGFUN Touch Probe & \href{https://a.co/d/f7wMJTQ}{Amazon} & 1 & \$76.99 \\
27 & MakerHawk 6pcs Optical Endstop & \href{https://a.co/d/0NPA5A1}{Amazon} & 1 & \$10.99 \\
28 & Carrier Drag Chain & \href{https://a.co/d/3KFZNA3}{Amazon} & 1 & \$13.89 \\
29 & Solenoid Air Valve & \href{https://a.co/d/8m1Z1gh}{Amazon} & 1 & \$9.99 \\
30 & Bowden Tube & \href{https://a.co/d/0gkupyu}{Amazon} & 1 & \$11.99 \\
31 & 0.4 mm Nozzle  & \href{https://a.co/d/65o17MI}{Amazon} & 1 & \$13.99 \\
32 & 3D Printer Extruder& \href{https://a.co/d/j0h7Tl4}{Amazon} & 1 & \$9.99 \\
33 & Plastic Tubing for Air (10 ft) & \href{https://www.mcmaster.com/5233K126/}{McMaster} & 1 & \$13.40 \\
34 & Hose Clamp (Pack of 10) & \href{https://www.mcmaster.com/5388K16/}{McMaster} & 1 & \$11.01 \\
35 & Barbed Hose Fitting 1/4 NPT & \href{https://www.mcmaster.com/5350k125/}{McMaster} & 1 & \$4.82 \\
36 & Barbed Hose Fitting 1/8 NPT & \href{https://www.mcmaster.com/5350K124/}{McMaster} & 2 & \$21.00 \\
37 & Air Regulator & \href{https://www.mcmaster.com/8812K52/}{McMaster} & 1 & \$30.96 \\
38 & Steel Ball Bearing & \href{https://www.mcmaster.com/6153K22/}{McMaster} & 10 & \$155.70 \\
39 & Flanged Steel Ball Bearing & \href{https://www.mcmaster.com/57155K628/}{McMaster} & 12 & \$217.08 \\
40 & Thrust Bearing & \href{https://www.mcmaster.com/5909K29/}{McMaster} & 6 & \$39.78 \\
41 & Low Profile Bearing & \href{https://www.mcmaster.com/6656K207/}{McMaster} & 2 & \$98.22 \\
42 & Rotary Shaft & \href{https://www.mcmaster.com/4143N17/}{McMaster} & 6 & \$64.02 \\
43 & M6 Flange Nut (Pack of 100) & \href{https://www.mcmaster.com/96194A101/}{McMaster} & 1 & \$7.41 \\
44 & M6 x 1mm, 35mm Screw (Pack of 10) & \href{https://www.mcmaster.com/91290a202/}{McMaster} & 1 & \$4.75 \\
45 & M4 x 0.7mm, 22mm Screw (Pack of 50) & \href{https://www.mcmaster.com/92095A479/}{McMaster} & 1 & \$6.89 \\
46 & M4 x 0.7mm, 16mm Screw (Pack of 100) & \href{https://www.mcmaster.com/92125A194/}{McMaster} & 1 & \$12.86 \\
47 & M8 x 1.25mm, 16mm Screw (Pack of 25) & \href{https://www.mcmaster.com/92125A282/}{McMaster} & 1 & \$16.20 \\
48 & M4 x 0.7mm, 22mm Screw (Pack of 100) & \href{https://www.mcmaster.com/92125A201/}{McMaster} & 1 & \$11.66 \\
49 & M3 x 0.5mm, 10mm Screw (Pack of 100) & \href{https://www.mcmaster.com/92125A130/}{McMaster} & 1 & \$7.31 \\
50 & Extension Spring (Pack of 6) & \href{https://www.mcmaster.com/9432K27/}{McMaster} & 1 & \$12.41 \\
51 & 3D Printing Hotend & \href{https://www.phaetus.com/en-us/products/rapido-hotend}{Phaetus} & 1 & \$79.00 \\

\midrule
\multicolumn{4}{r}{Total} & \$7{,}281.27 \\
\end{longtable}

\subsection*{Supplementary Movies}

\textbf{Movie S1.} A Micro-scale FPM. \\
\textbf{Movie S2.} A Centimeter-scale FPM. \\
\textbf{Movie S3.} A Meter-scale FPM.\\
\textbf{Movie S4.} Manufacturing an FPM Without Measurement. \\
\textbf{Movie S5.} Robotic FPM: Application in Metrology. \\
\textbf{Movie S6.} Robotic FPM: 3D Printing in a constrained workspace. \\

\end{document}